\documentclass[11pt]{article}

\usepackage{acl} 
\usepackage{times}
\usepackage{latexsym}
\usepackage[T1]{fontenc}
\usepackage[utf8]{inputenc}
\usepackage{microtype}

\usepackage{booktabs}
\usepackage{array}
\usepackage{multirow}
\usepackage{graphicx}
\usepackage{amsmath}
\usepackage{amssymb}
\usepackage{algorithm}
\usepackage{algpseudocode}
\usepackage{xcolor}
\usepackage{enumitem}
\usepackage{float}
\usepackage{placeins}
\usepackage{url}
\usepackage{hyperref}
\usepackage{tikz}
\usetikzlibrary{arrows.meta,positioning,shapes.geometric,patterns}

\hypersetup{
  pdftitle={},
  pdfauthor={},
  pdfsubject={},
  pdfkeywords={},
  pdfcreator={},
  pdfproducer={}
}

\newcommand{\GA}{\textsc{GA}}
\newcommand{\EA}{\textsc{EA}}

\title{Taxonomy-Targeted Error Generation for Quantitative Reasoning}

\author{Xinming Yang \\
  CUNY Graduate Center
  \And
  Jun Li \\
  CUNY Queens College \& CUNY Graduate Center}

\begin{document}
\maketitle

\begin{abstract}
Personalized tutoring, teacher preparation, and education research can
benefit from worked errors annotated by the mechanisms that produced
them. Authentic student errors with such cognitive labels are costly to
collect and share, motivating the study of whether LLMs can generate
taxonomy-targeted synthetic errors as complementary candidate material.
We present a task-specific framework that generates errors targeted to a
five-class Bloom-informed student-error taxonomy. A Generation Agent (\GA{})
drafts a candidate erroneous solution conditioned on a target class, and an
Examination Agent (\EA{}) judges whether the draft is incorrect and class-consistent.
The framework yields a reusable recipe for building class-stratified
synthetic error datasets where authentic student corpora are unavailable. As
a secondary diagnostic, targeted error generation is substantially harder
than free-form incorrect-answer generation, and answer-grounding
contributes more than expanded examples or external textbook content.

\end{abstract}

\section{Introduction}
\label{sec:intro}

Educational applications can benefit from worked errors labeled
by their underlying mechanisms. Personalized tutoring systems can use
such examples as candidate material for exercises and feedback calibrated
to specific learner weaknesses~\citep{vanlehn2011tutoring}.
Teacher-preparation programs use catalogs of common misconceptions
to train instructors to recognize them in student work, and
empirical evidence indicates that teachers' ability to identify
likely student misconceptions is a stronger predictor of classroom
learning gains than knowledge of the correct answer
alone~\citep{sadler2013teachers}. Education researchers studying
how misconceptions form, persist, and are repaired need
controlled collections of errors stratified by cognitive
type~\citep{smith1993misconceptions,anderson2001taxonomy}.
Exam-item designers likewise need distractor options that correspond to
plausible student reasoning rather than arbitrary wrong answers, a
property without which standard psychometric models systematically
misestimate student understanding~\citep{sadler1998psychometric}.

The bottleneck shared by these uses is access to suitably annotated data.
Collecting authentic student errors requires recruitment, consent, and
expert annotation; sharing them further requires de-identification and
data-governance safeguards. Learning-analytics work identifies consent,
privacy, de-identification, governance, and disclosure as barriers to
using student records at scale
\citep{slade2013learning,pardo2014ethical,usdoe2026ferpa}. Surveys of open
learning-analytics datasets also find uneven coverage across subject
areas \citep{svabensky2026open}. Although these requirements do not make
small-scale student studies infeasible, they increase the cost of building and
releasing labeled corpora at scale. More importantly, even
available corpora rarely annotate the cognitive mechanism
behind a wrong answer, {\em i.e.}, whether it reflects a lapse of
attention, a missing concept, an inappropriate solution procedure,
or a structural misframing
\citep{anderson2001taxonomy,brown1978diagnostic,vanlehn1990mind}.

\emph{Taxonomy-targeted errors} may complement authentic data by
supplying candidate material for these applications, analogous to prior uses
of privacy-preserving synthetic educational data \citep{vie2022privacy}.
We study taxonomy controllability: whether a generator can produce responses
that are
(i)~genuinely wrong, (ii)~consistent with a \emph{specified}
cognitive failure mode rather than arbitrarily incorrect, and
(iii)~obtainable under a single protocol that applies unchanged across
quantitative subjects.
These three criteria define our objective: scalable, guided generation of
incorrect worked answers conditioned on a requested error class. A usable
response must also commit to a concrete answer: refusals that plead
ignorance or uncertainty count as failures in our success metric
(\S\ref{sec:metric}).

We operationalize this controlled generation with a
task-specific \GA/\EA{} architecture. Its target labels derive from the
seven-class student-error taxonomy introduced by
\citet{zehetmeier2015classification}, which was itself organized with
reference to the cognitive dimensions of revised Bloom's
taxonomy~\citep{anderson2001taxonomy}, and we detail our
five-class adaptation in Section~\ref{sec:taxonomy}. Conditioned on a target
label, the \GA{} drafts a solution, and the \EA{} judges whether it is incorrect
and class-consistent. We then vary three design choices, {\em i.e.}, whether the
\GA{} sees the correct solution, whether failed attempts are fed
back into the next generation, and whether class judgment is done by
a prompted model or by a classifier fine-tuned on post-hoc human
corrections.
The result is a reusable recipe for building class-stratified error
datasets in subjects where authentic student corpora are scarce.

\textbf{Contributions.} Methodologically, we define a cross-subject,
open-ended controlled-generation task and protocol, with success requiring a
worked response to be incorrect, non-refusing, and target-class consistent;
we release the framework with its prompted- and fine-tuned-\EA{} variants
and an 1{,}800-record two-backend replication
artifact.\footnote{\url{https://github.com/junli-cuny/taxonomy-targeted-error-generation}.} Empirically, a
2{,}700-cell pilot over nine conditions and three backends shows that
targeted error generation is substantially harder than free-form
incorrect-answer generation; that success
can generally be raised over single-pass generation by grounding the \GA{}
in the gold answer and adding \EA{} filtering with rejection feedback,
although the size---and sometimes the direction---of each component's
contribution depends on the backend; and that many remaining pairwise
differences fall within question-cluster resampling variation.
The targeted-error success rate thus exposes
model behavior that correctness-only benchmarks miss.
Appendix~\ref{sec:related} reviews related work on reasoning
evaluation, educational error analysis and generation, LLM judges,
and synthetic error datasets.

\section{Error Taxonomy}
\label{sec:taxonomy}

\citet{zehetmeier2015classification} developed a seven-class student-error
taxonomy for programming education, organized with reference to the
cognitive dimensions of the revised Bloom's taxonomy
\citep{anderson2001taxonomy}: \textsc{remember},
\textsc{understand}, \textsc{apply}, \textsc{analyze}, \textsc{evaluate},
and \textsc{create}. Their seven original classes are
\textit{mental typo}, \textit{knowledge gap}, \textit{misconception},
\textit{wrong choice}, \textit{structural blindness}, \textit{quality
gap}, and \textit{lack of innovation}. Our adaptation retains the five
classes that describe incorrect responses and excludes \textit{quality
gap} and \textit{lack of innovation} because they describe suboptimal but
still essentially correct responses. The retained five classes are
summarized in Table~\ref{tab:taxonomy}.

\begin{table}[t]
\centering
\small
\setlength{\tabcolsep}{4pt}
\begin{tabular}{p{0.06\columnwidth}p{0.25\columnwidth}p{0.54\columnwidth}}
\toprule
\textbf{ID} & \textbf{Class} & \textbf{Description} \\
\midrule
E1 & Mental typo / sloppy work & Arithmetic or transcription slip in an otherwise correct trajectory. \\
E2 & Knowledge gap & Missing definition, term, or formula. \\
E3 & Misconception & Faulty mental model fitted to prior experience, or absent concept. \\
E4 & Wrong choice & Wrong problem classification or wrong solution procedure selected. \\
E5 & Structural blindness & Failure to distinguish components and their interaction in the given setting. \\
\bottomrule
\end{tabular}
\caption{Five-class student error taxonomy used throughout. Our adaptation
of the seven-class scheme from \citet{zehetmeier2015classification}
retains its five error-producing classes. We exclude \textit{quality gap}
and \textit{lack of innovation} because they describe \emph{suboptimal
correct} rather than \emph{incorrect} responses.}
\label{tab:taxonomy}
\end{table}

This taxonomy gives us a controlled vocabulary for the rest of the
paper. An \emph{error generation} task is now a 3-tuple
\textsf{(question, target class, generated response)}, and an
evaluator can
ask whether the response is incorrect and whether it falls in the
target class.

\section{Task and \GA/\EA{} Implementation}
\label{sec:framework}

Recall from \S\ref{sec:intro} that a practical synthetic-error
generator must produce responses that are (i)~genuinely wrong,
(ii)~consistent with the specified target class, and
(iii)~obtainable under a common protocol across subjects. The first
two requirements are hard to satisfy jointly: a model may simply
solve the problem correctly, and when it does err, the realized
error may fall in a class other than the one requested. To make
this concrete, we ran a small pilot in which GPT-5 was asked to
produce a target-class wrong answer for each of 20 questions drawn
from TheoremQA~\citep{chen2023theoremqa}, a theorem-driven QA dataset
of domain-expert-curated science problems covering math, physics,
EE/CS, and finance, paired with each of five target classes
($N=100$ question--target-class cells), with the gold answer and
class definitions supplied as grounding. On author inspection of the
resulting drafts, GPT-5 fails to match the requested class on $18\%$
of all 100 cells, and on $50\%$ of the 20 cells whose target is
structural blindness. We also ran this workflow in which we use GPT-5 instead of
a human run as a judge, and the observed mismatch rates were $8\%$ overall and
$20\%$ on structural-blindness cells.

\begin{figure}[t]
  \centering
  \begin{tikzpicture}[
    node distance=7mm and 4mm,
    every node/.style={font=\small},
    proc/.style={draw, rounded corners=2pt, align=center, inner sep=4pt,
                 minimum width=4cm, minimum height=7mm},
    io/.style={draw, trapezium, trapezium left angle=75, trapezium right angle=105,
               align=center, inner sep=4pt, minimum width=4cm, minimum height=7mm},
    data/.style={draw, rounded corners=10pt, align=center, inner sep=4pt,
                 minimum width=4cm, minimum height=7mm, fill=black!3},
    arrow/.style={-{Latex[length=2mm]}, thick}
  ]
    \node[io] (q) {Question $+$ target error class};
    \node[proc, below=of q] (ga) {Generation Agent (\GA)};
    \node[data, below=of ga] (cand) {Candidate erroneous response};
    \node[proc, below=of cand] (ea) {Examination Agent (\EA)};
    \node[io, below=of ea, fill=black!5] (out) {Stored final response};

    \draw[arrow] (q) -- (ga);
    \draw[arrow] (ga) -- (cand);
    \draw[arrow] (cand) -- (ea);
    \draw[arrow] (ea) -- node[right, font=\scriptsize]{Accept} (out);

    \draw[arrow] (ea.west) -- ++(-1.0,0)
                |- (ga.west)
                node[pos=0.25, rotate=90, above, font=\scriptsize]{EA reject; regenerate};
    \draw[arrow] (ea.east) -- ++(1.0,0)
                |- (out.east)
                node[pos=0.25, rotate=90, above, font=\scriptsize]{Cap reached};
  \end{tikzpicture}
  \caption{The \GA/\EA{} loop. The Generation Agent drafts a candidate
  incorrect response for a target error class; the Examination Agent
  judges whether the response is incorrect and matches the requested
  class. On \EA{} rejection, the \GA{} regenerates a candidate while budget
  remains. A capped
  run stores its last candidate even if the \EA{} rejects it;
  conditions P0/P2 bypass the \EA{}
  (Algorithm~\ref{alg:generation-scoring}).}
  \label{fig:pipeline}
\end{figure}
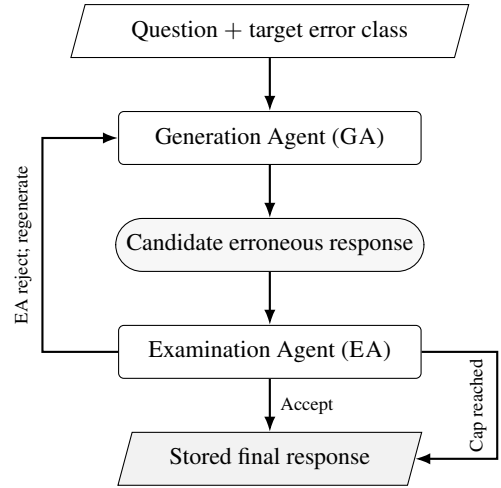

Motivated by the above pilot,\footnote{This single-backend, author-inspected check was not a
matched component comparison, so we use it only as motivation. The
evidence about single-pass generation and \EA{} use comes from the
full comparisons in \S\ref{sec:tier1-pipelines}.}
 we decouple drafting from judging into
two pre-trained agents: a Generation Agent (\GA)
is optimized for productive error construction, and an Examination Agent (\EA)
is optimized for class-conditional filtering. Given a question $q$ and target
error class $c \in \{1,\ldots,5\}$, the \GA{} produces a candidate and, when
the condition includes an \EA{}, that agent determines whether another draft
is required. Figure~\ref{fig:pipeline} sketches the loop. The \EA{} rejects
drafts that drift into neighboring classes. On \EA{} rejection, the \GA{}
regenerates a candidate up to the run's attempt budget (\S\ref{sec:main}).
The \GA{} and \EA{} configurations vary across nine pipeline
conditions (P0--P8), which are elaborated in \S\ref{sec:pipelines}. 

This architecture is subject-agnostic by design: the \GA{} and \EA{}
consume only the question text and the class definitions, with no
subject-specific module, so the same configuration and prompts
transfer across subjects unchanged. Requirement~(iii) is therefore
satisfied at the design level; the realized cross-subject rate is not
invariant in practice and is reported in Appendix~\ref{app:subject}.

\subsection{Generation Agent}
The Generation Agent (\GA) receives a question $q$ together with
the five class definitions, and is asked to produce a response $r$
that exemplifies a specified target class
$c \in \{1,\ldots,5\}$. We deliberately use a
\emph{single} multi-class \GA{} rather than five class-specialist
agents. In preliminary experiments, class-specialist agents produced
outputs whose realized error class often drifted into a sibling
class (especially $3{\leftrightarrow}4$), because each specialist
lacked the cross-class context needed to distinguish its target from
adjacent failure modes.

\subsection{Examination Agent}
The prompted Examination Agent (\EA) receives the question, gold
answer, target class, and generated response, and outputs a binary
judgment along two axes: whether the response is actually incorrect,
and whether it matches the target class. It accepts only when both
hold; otherwise it returns a rejection and justification. Its user
message supplies the same Q--A--E (question, erroneous answer,
explanation) exemplar block used by the \GA{}.

For lighter-weight scoring at scale, we additionally fine-tune a
BERT-base-uncased \citep{devlin2019bert} classifier as an alternative
\EA{} backend (P8), trained on $1{,}600$ author-annotated
\textsf{(3-tuple, judgment)} examples balanced across the five
classes. Unlike the prompted \EA{}, it is conditioned on the requested
class: its input serializes the question, response, and target
class, and its output is a single binary judgment --- valid targeted
error or not --- rather than a predicted class
(Appendix~\ref{app:annotation}). The \GA{} itself is never fine-tuned
in any pipeline; this keeps the framework a measurement instrument
whose numbers reflect the underlying LLM's off-the-shelf capacity
rather than our training data, and lets any pre-trained LLM serve as
a drop-in substitute.

\begin{table*}[t]
\centering
\footnotesize
\begin{tabular}{lp{0.4\textwidth}p{0.4\textwidth}}
\toprule
\textbf{Pipeline} & \textbf{\GA{} information} & \textbf{\EA{} role} \\
\midrule
P0 (baseline) & Base prompt; no correct answer & --- \\
P1 & Base + correct answer; unchanged after rejection & Base block \\
P2 & As P1 & --- (no \EA{}) \\
P3 & Base + correct answer + latest failure & Base block \\
P4 & As P3 & Base block + context \\
P5 & Class-dependent hybrid + latest failure & Class-dependent hybrid \\
P6 & Expanded + correct answer + latest failure & Expanded block \\
P7 & As P6 + context & Expanded block + context \\
P8 & As P1 & Fine-tuned classifier \\
\bottomrule
\end{tabular}
\caption{The nine evaluated conditions. Base has eight examples
($2/2/2/1/1$ for E1--E5); expanded has 25 (five per class). ``Context''
is a stored LLM-generated subject summary, not retrieved textbook text.
Exact context recipients are in Appendix Table~\ref{tab:context-recipients}.
For P1 and P3--P8, on \EA{} rejection the \GA{} regenerates a candidate.
P8 replaces the prompted \EA{} with the archived fine-tuned classifier.}
\label{tab:pipelines}
\end{table*}

\subsection{Pipeline Conditions}
\label{sec:pipelines}
We evaluate nine pipeline conditions, comprising one baseline (P0) and eight
variants (P1--P8; Table~\ref{tab:pipelines}). The baseline gives \GA{} only the
question, target error type, and few-shot examples, without the correct answer.
All eight variants P1--P8 are \emph{answer-grounded}: they expose the gold final
answer to \GA{}, but none exposes a gold reasoning trace. P1--P7 vary along
three further axes. First, this answer grounding tests whether
knowing the right solution helps it construct a \emph{targeted} wrong
one rather than an arbitrary mistake. Second, some variants add
richer conditioning, such as expanded few-shot examples or
LLM-generated auxiliary subject context, testing whether additional
scaffolding helps shape the output. 
Third, every condition with an \EA{}
(P1, P3--P8) follows the same rule: on \EA{} rejection, the \GA{}
regenerates a candidate. P3--P7
additionally expose the latest failure to the \GA{}, whereas P1 and
P8 regenerate from unchanged \GA{} input. P8 replaces the
prompted \EA{} with the fine-tuned BERT-base-uncased classifier
described above, making it the only pipeline that incorporates
human-supervised training.

\section{Experiments}
\label{sec:experiments}

\subsection{Evaluation Metric}
\label{sec:metric}
We separate answer correctness, realized error class, and refusal
status. Let $C_{q,c}=1$ iff the final response is correct,
$R_{q,c}$ be its realized class, and $F_{q,c}=1$ iff it is a
refusal. Refusals take priority and leave $C,R$ as N/A. Otherwise
the terminal outcome is \textsf{CorrectAns--ClassNA} when $C=1$,
\textsf{WrongAns--TargetClass} when $C=0,R=c$, and
\textsf{WrongAns--OtherClass} when $C=0,R\ne c$.
For question--target-class cells $\mathcal{D}$, the success indicator
and \textbf{targeted-error generation success rate (TEGSR)} are

{\small
\begin{equation}
  \begin{aligned}
  S_{q,c}&=\mathbf{1}[C_{q,c}=0]\mathbf{1}[R_{q,c}=c]
             \mathbf{1}[F_{q,c}=0],\\
  \operatorname{TEGSR}(\mathcal{D})
           &=\frac{1}{|\mathcal{D}|}
             \sum_{(q,c)\in\mathcal{D}}S_{q,c}.
  \end{aligned}
  \label{eq:tegsr}
\end{equation}
}

\noindent The \EA{}'s accept/reject decision controls only whether the \GA{}
resamples; it does not enter Equation~\ref{eq:tegsr}, which scores
the final stored response. Operationally, the \EA{} accepts
a draft when its extracted class matches the target $c$.
Algorithm~\ref{alg:generation-scoring} (Appendix~\ref{app:examples})
gives the complete loop.

\paragraph{Refusal status.} A usable targeted error must be a concrete
wrong answer. Explicit epistemic refusals ({\em e.g.}, ``I'm not sure'')
receive the distinct outcome \textsf{Refusal} and $S=0$; Appendix~
\ref{app:refusal} gives the regex and released \textsf{is\_refusal} flag.

Process- and cost-side metrics (retry count, token consumption,
\$-cost, latency, verifier agreement) are reported alongside
the appendices that use them. We compare these auxiliary metrics
in Appendix~\ref{app:redo} (retry counts),
Appendix~\ref{app:ablation-cascade} (verifier agreement), and
Appendix~\ref{app:compute} (token consumption, \$-cost, and
latency).

\subsection{Main Phase}
\label{sec:main}
The main evaluation draws up to 200 questions from
TheoremQA~\citep{chen2023theoremqa}, spanning combinatorics,
numerical analysis, graph algorithms, real and complex analysis,
physics, finance, optics, and computer networks. Two tiers with
distinct roles use different subsets of this pool: Tier-1
(\emph{Provisional Pilot}) sweeps the first 20 questions under full
outcome labeling
(\S\ref{sec:tier1-base}--\ref{sec:tier1-backend}), while Tier-2
(\emph{Operational Scale Run}) runs GPT-5 pipelines label-free over
the remaining 180 (\S\ref{sec:tier2}).

Tier-1 crosses its 20 questions with five
classes, nine conditions, and three backends (2{,}700 cells). In particular, we allow its
\EA{} loops to keep retrying until acceptance. This
first-20 convenience sweep was limited by single-author-verification
cost. The realized 6/7/3/4 question split covers
combinatorics/graph, analysis, linear algebra, and physics/probability.
Its submitted sheets retain only the legacy joint label
$h_{q,c}=1$ iff an author judged the response incorrect and
target-class consistent, plus $F_{q,c}$. Consequently, every
reported score is recovered exactly as $S_{q,c}=h_{q,c}(1-F_{q,c})$,
but a non-refusal $h=0$ cannot be split retrospectively between the
two non-target outcomes. We leave those decomposed release fields
null rather than fabricate labels (Appendix~\ref{app:dataset}).

Tier-2, by contrast, caps each cell at five total attempts and
stores the fifth candidate even if rejected. It is a larger run on
GPT-5 covering P1, P3, and P8 over 180 additional questions.
P1 and P3 were chosen as they dominate other pipelines on the GPT-5
deployment backend in Tier-1 (Table~\ref{tab:overall}, $0.70/0.69$). P8 is selected on architectural grounds (a
local BERT classifier vs.\ an API round-trip) to test whether its
near-single-pass acceptance persists outside the first-20 pilot.
Due to its scale, Tier-2 is label-free. It reports only operational quantities
--- \EA{} acceptance, retries, cost, and latency (\S\ref{sec:tier2};
cost details in Appendix~\ref{app:compute}), and provides neither
TEGSR nor independent pipeline validation.

\subsection{Models}
We evaluate \GA{} and \EA{} with three backends. These are a
mixed OpenAI o3 / GPT-4o configuration (\GA{} = OpenAI o3, \EA{} = GPT-4o),
GPT-5, and GPT-5-mini. We use the OpenAI API's system prompt slot for the
class definitions across all configurations. Section~\ref{sec:results}
reports per-model aggregates. Full per-cell breakdowns are in the appendix.

\subsection{Prompting Protocol}
For each \textsf{(question, target class)} pair, the \GA{} prompt
assembles three parts: the five class definitions, a Q--A--E exemplar
block, and the condition-specific inputs of Table~\ref{tab:pipelines}
(gold answer, latest rejected draft, generated auxiliary context).
Two exemplar inventories exist: P0--P4 and P8 use the base block,
P6/P7 use the expanded block, and P5 mixes them (base for E1/E2/E5,
expanded for E3/E4); the prompted \EA{} receives the block matching
its condition. Both inventories were manually curated by the authors
and are held fixed across backends, so every reported \GA{} condition
is few-shot and all backends see identical prompts. Exact prompts and block contents appear in Appendix~\ref{app:prompts}.

\section{Results}
\label{sec:results}

\begin{table}[t]
\centering
\small
\setlength{\tabcolsep}{5pt}
\begin{tabular}{llccc}
\toprule
\textbf{Pipeline} & \textbf{Description} & \textbf{o3+4o} & \textbf{GPT-5} & \textbf{mini} \\
\midrule
P0 & baseline & 0.55 & 0.63 & 0.58 \\
P1 & answer + EA & 0.50 & \textbf{0.70} & 0.69 \\
P2 & answer, no EA & 0.62 & 0.62 & 0.63 \\
P3 & P1 + feedback & \textbf{0.71} & 0.69 & 0.66 \\
P4 & EA context & 0.66 & 0.68 & 0.58 \\
P5 & hybrid context & 0.66 & 0.61 & 0.69 \\
P6 & expanded examples & 0.62 & 0.61 & 0.61 \\
P7 & GA+EA context & 0.70 & 0.62 & \textbf{0.70} \\
P8 & fine-tuned \EA{} & 0.65 & 0.66 & \textbf{0.70} \\
\bottomrule
\end{tabular}
\caption{Provisional Tier-1 TEGSR on the first-20 convenience sweep; \EA{} loops are uncapped. Exact question-cluster intervals are in Appendix~\ref{app:tier1-cluster-intervals} and per-class results in Appendix~\ref{app:per-pipeline}. ``mini'' = GPT-5-mini.}
\label{tab:overall}
\end{table}

\subsection{Targeted Error Generation Is Hard}
\label{sec:tier1-base}
Tier-1 is the paper's provisional TEGSR pilot. It applies all nine conditions
and three backends to the first 20 questions. \EA{} conditions retry until
acceptance; P0/P2 store one draft. One unblinded author, aware of target
and condition, assigned all joint labels in one post-storage pass. 

Producing a controllable wrong answer is not automatic even
when the question and target class are explicitly supplied. The
naive baseline (P0) leaves $37$--$45$ percentage points below
the ceiling on every backend (Table~\ref{tab:overall}). Under the
conservative non-refusal criterion, E2 (knowledge gap) is the dominant
bottleneck: its P0 rates are $0.10/0.20/0.15$, and across all conditions
it ranges from $0.00$ to $0.35$. Every one of the 424 regex-flagged
Tier-1 responses targets E2. This does not show that the models cannot
produce knowledge-gap-associated text; rather, many such responses use
explicit epistemic disclaimers that our operational criterion excludes.
E5 remains a secondary difficulty ($0.30$--$0.75$ across conditions).
Unlike E2, this difficulty is not a refusal artifact: no E5 response
is regex-flagged, so E5 misses fail the joint
incorrectness-and-target-class criterion itself. The same class is
costly upstream: in the \S\ref{sec:framework} pilot, half of the
grounded GPT-5 drafts targeting structural blindness missed the
requested class, and the \EA{} loops also work hardest there --- E5
has the highest mean retry count in five of the seven GPT-5 \EA{}
conditions (Table~\ref{tab:retries}), and the deep o3$+$GPT-4o retry
tail concentrates on E5 and E2 (Appendix~\ref{app:redo}). Which
class a failed E5 cell realizes instead cannot be recovered from the
submitted joint labels (\S\ref{sec:discussion}).
The remaining classes are reliably attainable: averaged over the
nine conditions, E1, E3, and E4 reach $0.77$--$0.92$ on every
backend, with individual condition--class cells as high as $1.00$;
on GPT-5, E1 and E3 lead (Figure~\ref{fig:perclass}), while E4 is
the top class on the other two backends.
Full per-class results are in Appendix~\ref{app:per-pipeline}.

\begin{figure*}[t]
  \centering
  \begin{tikzpicture}
    \begin{scope}[xscale=1.26, yscale=1.44]
      \draw[->] (0,0) -- (10.3,0) node[right]{};
      \draw[->] (0,0) -- (0,1.05) node[above, font=\scriptsize]{rate};

      \foreach \y in {0.25, 0.5, 0.75, 1.0}{
        \draw (0,\y) -- (-0.06,\y) node[left, font=\scriptsize]{\y};
      }
      \draw[dashed, gray!50] (0,1.0) -- (10.3,1.0);

      \def\pipelinedata{%
        {1/0.95/0.10/1.00/0.90/0.55/P1},%
        {2/0.90/0.00/0.95/0.75/0.50/P2},%
        {3/0.95/0.00/0.90/0.90/0.70/P3},%
        {4/0.85/0.05/1.00/0.95/0.55/P4},%
        {5/0.90/0.15/0.85/0.80/0.35/P5},%
        {6/0.90/0.05/0.80/0.85/0.45/P6},%
        {7/0.85/0.10/0.90/0.80/0.45/P7},%
        {8/0.85/0.15/0.85/0.85/0.60/P8}%
      }

      \foreach \i/\a/\b/\c/\d/\e/\lbl in \pipelinedata{
        \pgfmathsetmacro{\x}{(\i-1)*1.25 + 0.2}
        \fill[blue!60] (\x+0.00, 0) rectangle (\x+0.18, \a);
        \fill[green!60!black] (\x+0.20, 0) rectangle (\x+0.38, \b);
        \fill[orange!75] (\x+0.40, 0) rectangle (\x+0.58, \c);
        \fill[red!60] (\x+0.60, 0) rectangle (\x+0.78, \d);
        \fill[violet!60] (\x+0.80, 0) rectangle (\x+0.98, \e);
        \node[font=\scriptsize, anchor=north] at (\x+0.5, -0.04) {\lbl};
      }

      \begin{scope}[shift={(0,1.25)}, font=\tiny]
        \foreach \k/\col/\lbl in {0/blue!60/E1,
                                  1/green!60!black/E2,
                                  2/orange!75/E3,
                                  3/red!60/E4,
                                  4/violet!60/E5}{
          \pgfmathsetmacro{\lx}{\k*2.05}
          \fill[\col] (\lx, 0) rectangle (\lx+0.40, 0.10);
          \node[anchor=west] at (\lx+0.45, 0.05) {\lbl};
        }
      \end{scope}
    \end{scope}
  \end{tikzpicture}
  \caption{Provisional Tier-1 pilot TEGSR by pipeline and error class
  on the first 20 TheoremQA questions, GPT-5 backend. Each bar is
  a per-class proportion over $n{=}20$ cells. Under the conservative
  refusal exclusion, E2 (green) is lowest ($0.00$--$0.15$); E5
  (violet) remains the next-lowest class on several configurations
  ($0.35$--$0.70$).}
  \label{fig:perclass}
\end{figure*}
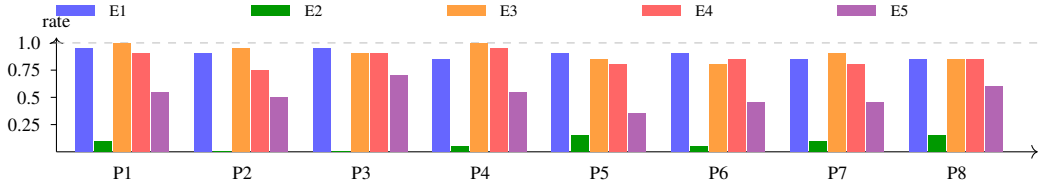

\subsection{Pipeline Comparison in the First-20 Pilot}
\label{sec:tier1-pipelines}
The pipeline-axis effects in Table~\ref{tab:overall} (per-class
breakdown on GPT-5 in Figure~\ref{fig:perclass}) identify which
design choices raise TEGSR and show that the size, and sometimes
the direction, of each gain depends on the backend.

\textbf{Answer grounding raises two of three backends.}
P0 and P2 are matched single-pass conditions without an \EA{}, and only
P2 supplies the gold final answer to the \GA{}. Their P0$\to$P2 rates
change from $0.55/0.63/0.58$ to $0.62/0.62/0.63$ (o3$+$GPT-4o /
GPT-5 / GPT-5-mini), or $+7/-1/+5$ pp. Grounding also reshapes
classes: the largest E2 change is backend-specific, with GPT-5
falling from $0.20$ to $0.00$ while GPT-5-mini rises from $0.15$ to
$0.35$. Across these six settings, $37\%$--$45\%$ of outputs still
miss the joint incorrectness-and-target-class criterion, which is the
headroom that the \EA{} loop targets next. 

\textbf{The \EA{} loop reaches the highest rates.} Adding
prompted-\EA{} filtering and resampling (P2$\to$P1,
$0.62/0.62/0.63\to0.50/0.70/0.69$) and then rejection feedback
(P1$\to$P3, $\to0.71/0.69/0.66$) nets $+9/+7/+3$ pp over grounded
single-pass generation on every backend, attaining the top of the
o3$+$GPT-4o and GPT-5 columns ($0.71$, $0.70$) and coming within one
point of GPT-5-mini's best. The decomposition is backend-dependent. 
Filtering alone lowers o3$+$GPT-4o ($-12$ pp) while raising the
other two ($+8/+6$), and feedback recovers o3$+$GPT-4o ($+21$) while
roughly preserving the rest ($-1/-3$). Hence, neither component alone
is a universal remedy, but their combination is what Tier-2 carries
forward as P1/P3 (\S\ref{sec:main}).

\textbf{Exact question-set sensitivity.} How much would
Table~\ref{tab:overall} change if the pilot had drawn a different
set of 20 questions? We answer by recomputing every rate while
redrawing the 20 questions with replacement; each redrawn question
keeps its five class outcomes, and paired pipelines are compared on
the same redraw. Each result below pairs the observed rate with the
middle $95\%$ of its recomputed values in brackets, written
rate~[lower, upper]; the distribution over all $20^{20}$ ordered
redraws is aggregated exactly by dynamic programming rather than
enumerated or approximated by simulation
(Appendix~\ref{app:tier1-cluster-intervals}). On GPT-5 the choice
of questions matters about as much as the choice of pipeline: P1
scores $0.70$ [$0.62$, $0.78$] and P3 scores $0.69$ [$0.61$,
$0.75$], with a same-redraw gap of only $-0.01$ [$-0.07$, $+0.05$];
a single range is nearly as wide as the entire $0.61$--$0.70$ GPT-5
column. Moreover, of the 18 head-to-head pipeline comparisons, only
two have ranges that stay on one side of zero --- the o3$+$GPT-4o
effects already seen above, where rejection feedback helps ($+0.21$
[$+0.08$, $+0.36$]) and the implemented expansion hurts ($-0.09$
[$-0.17$, $-0.01$]). The other 16 could go either way, so among the
leading conditions one may choose on cost and latency rather than
rate, as the Tier-2 deployment does (\S\ref{sec:tier2}).
Appendix~\ref{app:tier1-cluster-intervals} reports all 27 rate
ranges and 18 comparisons. 

Because P2$\to$P1 changes two things at once --- drafts are both
judged and, when rejected, regenerated --- we also ask what the
\EA{}'s judging alone contributes. Each backend's prompted \EA{}
re-judged its own 100 stored P2 drafts exactly once, with nothing
regenerated and no new human labels; its accept/reject decisions
were then scored against the existing author labels. The result is a
trade-off between how much an \EA{} keeps and how good the kept
drafts are. The o3$+$GPT-4o \EA{} is strict: it keeps only 38 of 100
drafts, of which $0.66$ are valid (versus $0.62$ of all drafts), so
just 25 valid drafts per 100 survive. The GPT-5 \EA{} keeps 92 of
100 and raises the valid share among kept drafts from $0.62$ to
$0.65$; GPT-5-mini also keeps 92 but slips from $0.63$ to $0.61$.
This fixed-output replay measures judging only; it does not estimate
the end-to-end effect of retries or feedback (full prompted and P8
results in Appendix~\ref{app:p2-ea-replay}).

\textbf{The base 8-example block suffices (P3$\to$P6).}
P3 uses the base 8-example blocks for both agents and P6 uses the
expanded 25-example blocks; answer grounding, rejection feedback, and
the absence of auxiliary context are otherwise fixed. Rates change from
$0.71/0.69/0.66$ to
$0.62/0.61/0.61$ (o3$+$GPT-4o / GPT-5 / GPT-5-mini), or
$-9/-8/-5$ pp: the shorter, cheaper prompt is the better-performing
one on every backend, and only the o3$+$GPT-4o exact paired
interval excludes zero
(Appendix~\ref{app:tier1-cluster-intervals}). 
Appendix~\ref{app:prompts}
documents and releases both the source literal and exact JSON-encoded runtime
strings.

\textbf{Generated auxiliary context helps selectively.}
Whether a stored subject summary helps depends on the prompt it is
added to. Added to the expanded-block pipeline (P6$\to$P7), it is
broadly useful: rates rise from $0.62/0.61/0.61$ to
$0.70/0.62/0.70$, with four of five classes improving on
o3$+$GPT-4o and GPT-5-mini and E2/E3 improving on GPT-5. Added to
the base-block pipeline (P3$\to$P4, where only the \EA{} receives
it), it does not help: rates change from $0.71/0.69/0.66$ to
$0.66/0.68/0.58$, every redraw range includes zero, and E5 drops by
$0.15$ on every backend. 

\textbf{Fine-tuned vs.\ prompted \EA{} (P8 vs.\ P1) is a deployment
trade-off.} P8 replaces P1's prompted \EA{} with a fine-tuned BERT
classifier, keeping every other component identical. The matched
P1-versus-P8 point-estimate difference is $+15$ pp on o3$+$GPT-4o
($0.50\!\to\!0.65$), $-4$ on GPT-5, and $+1$ on GPT-5-mini. The two \EA{}s also
differ structurally. The prompted \EA{}'s acceptance threshold is
tunable per call via its system prompt, whereas P8's is fixed in
the classifier weights. P8 is cheaper at inference (single forward
pass, zero API tokens; \S\ref{sec:tier2}) but less adaptable;
framing P8 vs.\ a prompted \EA{} as an API-cost vs.\ on-device
trade-off is at least as informative as framing it as a quality
contest. Because the archived P8 training input contains its label, its
rates remain exploratory.

\subsection{Backend Comparison}
\label{sec:tier1-backend}
How large the between-backend gap is depends on the refusal rule.
Under the legacy author-judged joint label alone, the gap is large
and asymmetric: P1--P7 averages are $0.71/0.83/0.79$
(o3$+$GPT-4o / GPT-5 / GPT-5-mini), and E2 is the separator, with
legacy P1 E2 at $0.55/1.00/0.95$. The conservative epistemic-refusal
exclusion erases the gap --- $0.64/0.65/0.65$ --- precisely because
it strikes the E2 cells on which GPT-5 and GPT-5-mini appear
strongest (\S\ref{sec:tier1-base}). Under the normalized metric the
backends reach their similar averages by different routes. Their class profiles differ ---
GPT-5 is highest on average for E1/E3, GPT-5-mini for E2, and
o3$+$GPT-4o for E5 --- and so does the component each profits from:
o3$+$GPT-4o gains from rejection feedback ($+21$ pp, one of the two
effects robust to question redraws in \S\ref{sec:tier1-pipelines}),
whereas GPT-5 and GPT-5-mini gain from filtering alone ($+8/+6$ pp)
with little further change from feedback ($-1/-3$ pp). 

\subsection{Retry Behavior and Cost}
\label{sec:tier2}
Retry counts surface a hidden cost dimension. Two configurations
can reach the same TEGSR while one spends
$2$--$3\times$ the \GA{} compute. On Tier-1, the backend-retry
gap is almost entirely E5: P1 averages $2.45$ attempts on
o3$+$GPT-4o vs.\ $1.28$ on GPT-5 (Appendix~\ref{app:redo},
Table~\ref{tab:redo}), with the per-class breakdown showing the
gap concentrated on E5. Weak
backends are not uniformly slower; they are unreliable
specifically on cells where the \GA{} re-generates near-clones
until one happens to land in the target class, so the retry
budget is spent on diversification rather than judging. P8
reshapes \emph{which} class drives the long tail. The prompted
pipelines spend retries on E5 (Table~\ref{tab:retries}: P1
E5~$=0.90$), but P8's largest retry pocket is on E3 ($0.90$),
not E5. Its training data over-weighted E5, so it accepts
structural-blindness drafts quickly but is unusually strict on
misconception drafts. The residual ``hard class'' shifts with
the \EA{} implementation because the BERT calibration is set
offline rather than by a per-call adaptable prompt.

\begin{table}[t]
\centering
\small
\setlength{\tabcolsep}{4pt}
\begin{tabular}{lccccc}
\toprule
\textbf{Pipe} & \textbf{E1} & \textbf{E2} & \textbf{E3} & \textbf{E4} & \textbf{E5} \\
\midrule
P1 & 0.10 & 0.10 & 0.05 & 0.25 & \textbf{0.90} \\
P3 & 0.10 & 0.00 & 0.05 & 0.15 & 0.35 \\
P4 & 0.10 & 0.00 & 0.10 & \textbf{0.65} & 0.20 \\
P5 & 0.05 & 0.00 & 0.20 & 0.10 & 0.40 \\
P6 & 0.00 & 0.15 & 0.05 & 0.20 & 0.40 \\
P7 & 0.10 & 0.00 & 0.00 & 0.15 & 0.25 \\
P8 & 0.25 & 0.00 & \textbf{0.90} & 0.40 & 0.15 \\
\bottomrule
\end{tabular}
\caption{Provisional Tier-1 pilot mean retries per cell by pipeline and error
class. Each cell is
the average over 20 (question, target class) instances of the
number of \EA{}-induced retries (retries~$=$ \GA{} attempts~$-$~1,
floor~$0$). A value of $0.00$ means every instance was accepted
on the first draft; a value of $0.90$ means an average of $0.90$
retries per instance, with boldface marking notable retry peaks
within the table. P2 is omitted because it
has no \EA{} loop. Source: Appendix~\ref{app:redo},
Table~\ref{tab:redo}.}
\label{tab:retries}
\end{table}

Tier-1's unlimited retries are impractical at scale. Its retry
distribution motivates Tier-2's five-attempt cap. Overall, $3.6\%$ of Tier-1
cells require more than five attempts ($\le 1\%$ on GPT-5;
Appendix~\ref{app:redo}). The uneven scopes and post-hoc pipeline
selection are defined in \S\ref{sec:main}; this section reports only
label-free operational quantities, with costs in
Appendix~\ref{app:compute}, not validity evidence.

Two patterns hold at scale. First, the prompted-\EA{} loops and
the BERT \EA{} differ in \emph{shape}, not just in absolute
acceptance (Table~\ref{tab:tier2-accept}). P1/P3 climb gradually
but remain below full acceptance at the cap, while P8 is
near-saturated at $k{=}1$. Relative to five attempts, a one-attempt
budget loses 24.1 points for P1, 33.7 for P3, and 5.4 for P8; a
three-attempt budget loses 4.7, 6.5, and 1.0 points, respectively.
The contrast between P1 and P3
also persists across the whole budget range rather than appearing
only at the cap. P3's observed acceptance is above P1 at every $k$,
which indicates
that feeding back the rejected draft and \EA{} rationale is associated
with faster \emph{EA acceptance} in this run. Because Tier-2 is
label-free, this does not establish a higher single-author-verified
TEGSR or any independently verified TEGSR.

Second, the cost tails differ (Table~\ref{tab:tier2-cost}). Prompted
P1/P3 compound \GA{} and \EA{} retries and reach p95 above 63k
tokens / 5~min per cell. P8's local classifier adds no API tokens,
so its cost is \GA{}-only and tightly concentrated.

\begin{table}[t]
\centering
\small
\setlength{\tabcolsep}{5pt}
\begin{tabular}{lccccc}
\toprule
\textbf{Pipe} & $k{=}1$ & $k{=}2$ & $k{=}3$ & $k{=}4$ & $k{=}5$ \\
\midrule
P1 & 0.547 & 0.684 & 0.741 & 0.770 & 0.788 \\
P3 & 0.552 & 0.750 & 0.824 & 0.868 & 0.889 \\
P8 & 0.942 & 0.976 & 0.986 & 0.992 & 0.996 \\
\bottomrule
\end{tabular}
\caption{Cumulative Tier-2 \EA{} acceptance by attempt budget $k$ on GPT-5. A
fifth rejected candidate is stored but is not counted as accepted.}
\label{tab:tier2-accept}
\end{table}

\begin{table}[t]
\centering
\small
\setlength{\tabcolsep}{4pt}
\begin{tabular}{lrrrr}
\toprule
 & \multicolumn{2}{c}{\textbf{Tokens / cell}} & \multicolumn{2}{c}{\textbf{Latency (s)}} \\
\cmidrule(lr){2-3}\cmidrule(lr){4-5}
\textbf{Pipe} & \textbf{med} & \textbf{p95} & \textbf{med} & \textbf{p95} \\
\midrule
P1 & 14{,}951 & 67{,}642 & 70.5 & 385.6 \\
P3 & 15{,}059 & 63{,}874 & 69.6 & 349.9 \\
P8 & \phantom{0}4{,}947 & 14{,}148 & 24.3 & \phantom{00}79.5 \\
\bottomrule
\end{tabular}
\caption{Tier-2 per-cell cost on GPT-5 (same question scopes as
Table~\ref{tab:tier2-accept}). P1/P3 include \GA{}
and prompted-\EA{} tokens across retries; P8 counts only \GA{}
tokens because its local BERT \EA{} adds no API tokens.}
\label{tab:tier2-cost}
\end{table}

\section{Analysis and Discussion}
\label{sec:discussion}

\paragraph{Hard-class patterns.}
Figure~\ref{fig:perclass} shows that E2 is hardest under the stated
non-refusal rule. This is partly definitional: knowledge-gap prompts often
elicit explicit ignorance or uncertainty, and the conservative heuristic
excludes such outputs even when they later guess. E5 remains difficult on
several configurations, but the submitted joint labels do not identify which
alternative class occurs when a cell fails. A quantitative confusion matrix
requires new realized-class annotation and remains future work.

\paragraph{Matching components to backends.} Each backend has a
component it profits from: o3$+$GPT-4o improves most when rejected
drafts and \EA{} justifications are fed back, GPT-5 and GPT-5-mini
improve under \EA{} filtering alone, and answer grounding mainly
lifts the two backends that are weaker single-pass. The fixed-output
replay tells the same story from the verifier side: one prompted-\EA{}
design acts as a strict, low-coverage filter on one backend and as a
near pass-through on another (Table~\ref{tab:p2-ea-comparison}).
Only the o3$+$GPT-4o feedback gain is robust to question redraws;
the rest are point-estimate-level
(Appendix~\ref{app:tier1-cluster-intervals}). This argues for a
small per-backend pilot rather than one configuration everywhere; no
tested component is a backend-independent default. We also stop short of a
``trajectory-control'' claim because no condition receives a gold
reasoning trace (\S\ref{sec:limitations}). TEGSR still adds
information invisible to correctness-only benchmarks.

\section{Conclusion}
\label{sec:conclusion}

We presented a framework for class-stratified synthetic error
generation where labeled student errors are scarce, in which a \GA{}
drafts an error aimed at a requested taxonomy class and an \EA{}
checks that the draft is genuinely wrong and on-class. We find that asking for a \emph{specific} kind of wrong answer
is much harder than asking for any wrong answer. Success varies
far more with pipeline configuration than with backend choice.
Controlled error generation is thus a capability of its own, one
that correctness-only benchmarks do not measure.


\section*{Limitations}
\label{sec:limitations}

\textbf{Scale and human labeling.} Tier-1 is a first-20 convenience
sweep, chosen to bound manual verification cost rather than by power
analysis, probability sampling, or subject stratification. Its 6/7/3/4
subject distribution is descriptive, not a balanced design. The full
pool contains $2{,}700$ single-author-verified
$\langle$\textsf{question, target class, response}$\rangle$ cells
(20 questions $\times$ 5 classes $\times$ 9 conditions $\times$ 3
backends), so its TEGSR estimates are provisional. Tier-2
(\S\ref{sec:tier2}) adds label-free operational measurements on uneven
question scopes, not a larger validity sample. The prompted, fine-tuned,
and advanced automatic judges have not met a validated human-level gate
and are not used as ground truth. Tier-2 therefore neither provides TEGSR
nor compensates for Tier-1's sampling limitations. Scaling outcome
estimates requires human annotation or a future automatic judge that
first passes such a gate.

\textbf{Rule-based refusal status.} Refusal status is assigned by a
deterministic, intentionally conservative phrase matcher rather than independent
human labels. Apostrophe normalization prevents typography-dependent results,
but the matcher's precision and recall have not been measured. All 424 Tier-1
matches are E2 and 376 override a positive legacy joint label, so E2 rates are
especially sensitive to this operational definition. The planned blinded
annotation records refusal separately and can validate this heuristic.


\textbf{No real-student baseline (separate validity question).} The core
evaluation measures guided generation. It tests whether a response is incorrect,
non-refusing, and consistent with the requested class. It does not test whether
humans can distinguish an output's source as human or AI, nor whether the
outputs match the frequency or surface-form distribution of spontaneous
student errors. Those are separate external-validity questions that we did not
test in this submission. The closest existing student-misconception
corpora \citep{sun2025error,otero2024misconceptions,oli2024gaps}
each cover different subjects and label schemes from ours, so
matching against them requires a label re-projection step we have
not done. Until that comparison exists, the outputs should be
read as \emph{taxonomy-targeted synthetic errors}, without claims of
source indistinguishability, distribution matching, or educational validity.

\textbf{No reasoning-trace ablation.} The paper uses
``answer-grounded generation'' for P1--P8 because each receives the gold final
answer; P0 does not. No pipeline conditions on the gold \emph{reasoning trace} (the
step-by-step derivation that produces the answer). A new natural
pipeline would supply the trace and let us cleanly distinguish
answer-only grounding from trajectory grounding; we did not run
this and the ``trajectory'' framing in the discussion is therefore
weaker than ``answer-grounding.''



\textbf{Provisional single-author labels.} One author assigned the
submitted Tier-1 joint labels in a single pass while aware of the target
class and experimental condition; the procedure was not blinded. There
was no independent second annotation pass or adjudication, so
inter-annotator agreement is unavailable and target-aware judgment may
affect the reported TEGSR. A prospective blinded 240-cell study uses the
same two locked human label files for human--human reliability and
advanced-agent validation; it was not part of the submitted experiment
and no agreement result from it is claimed here.

\section*{Ethical considerations}
\label{sec:ethics}

This work studies the generation of taxonomy-targeted synthetic
incorrect answers. The proposed educational use is educator-reviewed candidate
material for error-diagnosis and correction exercises. We have not tested
learner outcomes or correspondence to spontaneous student errors, and the
outputs are not intended for deployment without educator review. Risks include
class-mismatched, misleading, or educationally unsuitable incorrect answers,
especially in subjects where our model coverage is thin. We mitigate the
within-taxonomy mismatch risk by (i)~tying generation to an explicit taxonomy,
(ii)~judging outputs with an \EA{}, and (iii)~publishing the fine-tuned \EA{}
for downstream verification; these measures do not close the separate
external-validity gap.

\paragraph{AI assistance disclosure.} We used AI-based coding and writing
assistants for software scaffolding, LaTeX formatting, and prose editing. All
experimental design, analysis, and claims are the authors' own; the commercial
LLMs evaluated in this paper are objects of study rather than writing aids.

\bibliography{custom}


\appendix
\section{Exact Question-Cluster Intervals}
\label{app:tier1-cluster-intervals}

The exact calculation aggregates the full discrete distribution rather
than drawing a finite Monte Carlo sample. Each resample is 20 i.i.d.
draws from the 20 question clusters, and every reported statistic is a
sum of per-draw integer contributions (a headline-rate total lies in
$0..100$; a paired-contrast total in $-100..100$), so its exact
distribution over all $20^{20}$ ordered resamples is the 20-fold
convolution of the single-draw distribution, computed by dynamic
programming. The calculation therefore needs neither a replicate
count nor a random seed, removes only Monte Carlo error, and makes no
additional model calls.

\begin{table}[t]
\centering
\scriptsize
\setlength{\tabcolsep}{2.5pt}
\begin{tabular}{lccc}
\toprule
\multicolumn{4}{c}{\textbf{(a) Headline TEGSR}} \\
\textbf{Pipe} & \textbf{o3+4o} & \textbf{GPT-5} & \textbf{mini} \\
\midrule
P0 & .55 [.46, .63] & .63 [.54, .71] & .58 [.50, .66] \\
P1 & .50 [.37, .62] & .70 [.62, .78] & .69 [.61, .77] \\
P2 & .62 [.54, .70] & .62 [.54, .69] & .63 [.51, .74] \\
P3 & .71 [.63, .78] & .69 [.61, .75] & .66 [.58, .74] \\
P4 & .66 [.58, .73] & .68 [.62, .74] & .58 [.50, .65] \\
P5 & .66 [.56, .76] & .61 [.52, .69] & .69 [.61, .77] \\
P6 & .62 [.53, .70] & .61 [.51, .69] & .61 [.51, .71] \\
P7 & .70 [.63, .76] & .62 [.51, .72] & .70 [.62, .77] \\
P8 & .65 [.55, .74] & .66 [.58, .74] & .70 [.63, .77] \\
\midrule
\multicolumn{4}{c}{\textbf{(b) Paired contrasts: second pipeline minus first}} \\
\textbf{Contrast} & \textbf{o3+4o} & \textbf{GPT-5} & \textbf{mini} \\
\midrule
P1$\to$P3 & +.21 [+.08, +.36] & -.01 [-.07, +.05] & -.03 [-.11, +.05] \\
P3$\to$P8 & -.06 [-.17, +.05] & -.03 [-.11, +.05] & +.04 [-.04, +.13] \\
P0$\to$P2 & +.07 [-.03, +.17] & -.01 [-.08, +.07] & +.05 [-.05, +.15] \\
P1$\to$P8 & +.15 [.00, +.31] & -.04 [-.11, +.02] & +.01 [-.06, +.09] \\
P3$\to$P4 & -.05 [-.12, +.01] & -.01 [-.07, +.05] & -.08 [-.16, .00] \\
P3$\to$P6 & -.09 [-.17, -.01] & -.08 [-.15, .00] & -.05 [-.14, +.04] \\
\bottomrule
\end{tabular}
\caption{Exact question-cluster bootstrap intervals for the provisional Tier-1 pilot. Each entry is the point estimate followed by an equal-tailed 95\% interval. The complete distribution over all $20^{20}$ ordered resamples of the 20 observed convenience-sampled questions is aggregated by dynamic programming; each selected question retains all five target classes. Panel (b) pairs pipelines within question. These intervals measure sensitivity to reweighting this observed question set, not uncertainty over all TheoremQA items or run-to-run generation variation. ``mini'' = GPT-5-mini.}
\label{tab:tier1-cluster-intervals}
\end{table}


\section{Generation, Selection, and Outcome Scoring}
\label{app:examples}

\begin{algorithm}[h]
\footnotesize
\caption{Generation, \EA{} selection, and post-hoc scoring for one
question--target-class cell.}
\label{alg:generation-scoring}
\begin{algorithmic}[1]
\Require question $q$, target class $c$, condition $p$, attempt budget $B$
\State $t\gets0$; $a\gets\mathrm{N/A}$; $u\gets\varnothing$
\If{$p\in\{\mathrm{P0},\mathrm{P2}\}$} \Comment{no \EA{}}
  \State $t\gets1$; $r^*\gets\Call{GA}{p,q,c,u}$
\Else
  \Repeat
    \State $t\gets t+1$; $r_t\gets\Call{GA}{p,q,c,u}$
    \State $(z_t,j_t)\gets\Call{EA}{p,q,c,r_t}$
    \If{$p=\mathrm{P8}$}
      \State $a\gets[z_t=1]$ \Comment{binary classifier}
    \Else
      \State $a\gets[z_t=c]$ \Comment{extracted prompted-\EA{} numeral}
    \EndIf
    \If{$\neg a$ and $p\in\{\mathrm{P3},\ldots,\mathrm{P7}\}$}
      \State $u\gets(r_t,j_t)$ \Comment{latest rejected draft and rationale}
    \EndIf
  \Until{$a$ or $t=B$}
  \State $r^*\gets r_t$
\EndIf
\State $\mathit{retries}\gets t-1$; $f\gets\Call{IsRefusal}{r^*}$
\If{\text{separate post-hoc labels are available}}
  \State $(C,R)\gets\Call{Annotate}{q,r^*}$ \Comment{correctness and realized class}
  \State $S\gets[ C=0 ][ R=c ][ f=0 ]$
  \State $o\gets\Call{TerminalOutcome}{C,R,f,c}$
\ElsIf{\text{legacy joint author label is available}}
  \State $h\gets\Call{HumanExamination}{q,c,r^*}$
  \Comment{$1$ iff incorrect and target-class consistent}
  \State $S\gets h(1-f)$
  \State $(C,R,o)\gets\Call{LegacyProjection}{h,f,c}$
\Else
  \State $C,R,h,o,S\gets\mathrm{N/A}$ \Comment{the scale run remains label-free}
\EndIf
\State \Return $(r^*,a,\mathit{retries},C,R,h,f,o,S)$
\end{algorithmic}
\end{algorithm}
\FloatBarrier

Here $B=\infty$ for Tier-1 conditions with an \EA{} and $B=5$ for
the capped scale runs. P1 and P8 regenerate from unchanged \GA{}
input, whereas P3--P7 receive only the latest rejected draft and
\EA{} rationale. In a capped run, termination with $a=0$ stores the
last rejected candidate; cap termination is therefore not \EA{}
acceptance. The submitted Tier-1 data use the legacy-label branch.
The legacy projection returns \textsf{Refusal} when $f=1$ and
\textsf{WrongAns--TargetClass} when $f=0,h=1$; when $f=0,h=0$ it
leaves $C,R,o$ as N/A because the source label does not distinguish
the other two outcomes. All post-hoc fields are computed after
selection and are never fed into regeneration. Thus an \EA{}-accepted
output may still have $S=0$, and a no-\EA{} output may have $S=1$.

\paragraph{Example~A. Correct output despite a class-1 request.}
\begin{quote}\small
\textbf{Q.} Fifteen more than a quarter of a number is 27. What is
the number?

\textbf{Requested class.} 1 (mental typo).

\textbf{Generated response.} A quarter of the number is $x/4$. So,
$x/4 + 15 = 27$. Subtract $15$: $x/4 = 12$. Multiply both sides
by~4: $x = 48$. \emph{Therefore, the number is~48.}

\textbf{Label.} \textsf{CorrectAns--ClassNA} -- the model answered
the problem correctly, so no error class applies.
\end{quote}

\paragraph{Example~B. \textsf{WrongAns--TargetClass} -- desired output.}
\begin{quote}\small
\textbf{Q.} Same as above. \textbf{Requested class.} 1 (mental
typo).

\textbf{Generated response.} $(1/4)\,n + 15 = 27$. Subtract~15:
$(1/4)\,n = 12$. Multiplying both sides by~3 to solve for~$n$:
$n = 12 \times 3 = 36$. So the number is~36.

\textbf{Label.} \textsf{WrongAns--TargetClass} -- an incorrect answer
caused by multiplying by~3 rather than~4, exactly the requested
mental-typo failure mode. An incorrect response assigned another
class would instead be \textsf{WrongAns--OtherClass}.
\end{quote}

\paragraph{Example~C. \textsf{Refusal}.} A response can read as
class-consistent in tone
yet fail to produce a concrete taxonomy-targeted error because it
admits not knowing. Such a response receives the distinct terminal
outcome \textsf{Refusal}.
\begin{quote}\small
\textbf{Q.} Is $W = \{[x, y] \in \mathbb{R}^2 : x \geq 0
\text{ and } y \geq 0\}$ a subspace of $\mathbb{R}^2$?
\textbf{Gold answer.} False. \textbf{Requested class.} 2
(knowledge gap).

\textbf{Generated response.} First, I know that to show a set is
a subspace of $\mathbb{R}^2$ we have to check certain properties,
but I'm not fully sure which ones are required. I vaguely recall
that it has something to do with being ``closed under addition''
and also something with ``scalar multiplication,'' yet I'm not
certain what ``closed'' really means in this context. Since I'm
unsure of the exact formal definition of a subspace, I can't
confidently decide if the set $W$ qualifies.

\textbf{Label.} \textsf{Refusal}. The response expresses ignorance
rather than committing to a concrete wrong
trajectory, so it is not a usable knowledge-gap exemplar even
though its tone superficially matches class~2. The automatic
\textsf{is\_refusal} flag fires on ``I'm not fully sure'',
``I vaguely recall'', and ``I can't confidently decide''.
\end{quote}

\subsection{One Question Across All Five Requested Classes}
\label{app:five-class-example}

The following final responses instantiate all five requested classes for one
TheoremQA item. The question is ``Compute
$\oint_{|z|=1}z^2\sin(1/z)\,dz=Ai$; what is $A$?'' The gold value is
$A=-\pi/3\approx-1.047$. To avoid fabricating or
relabeling examples, we selected existing Tier-1 records for this question
whose legacy joint label is 1 and refusal flag is 0. The records come from
different conditions and serve only as taxonomy illustrations, not a matched
pipeline comparison; their labels remain provisional under the disclosure in
Appendix~\ref{app:annotation}. The displays are lightly typeset and condensed;
the exact responses are recoverable from q06, the stated condition, backend,
and class in the released JSONL.

\paragraph{E1 (mental typo; P5, GPT-5-mini).}
\begin{quote}\small
Expand
$\sin(1/z)=\sum_{n=0}^{\infty}(-1)^n z^{-(2n+1)}/(2n+1)!$ and multiply by
$z^2$. The $z^{-1}$ term has $n=1$ and coefficient $-1/3!=-1/6$.
Thus the integral is $2\pi i(-1/6)=-\pi i/6$, so $A=-\pi/6$.
\end{quote}
The final simplification is a localized arithmetic slip. The value $2\pi/6$ is
$\pi/3$, not $\pi/6$.

\paragraph{E2 (knowledge gap; P2, GPT-5-mini).}
\begin{quote}\small
Since $\sin w$ is entire, treat $\sin(1/z)$ as a composition of an entire
function with $1/z$ and assume it remains analytic on and inside
$|z|=1$. Then $z^2\sin(1/z)$ is analytic, so Cauchy's theorem gives an
integral of 0 and hence $A=0$.
\end{quote}
The response lacks the needed fact that composition with $1/z$ introduces a
singularity at zero; multiplying by $z^2$ does not remove the full Laurent
tail.

\paragraph{E3 (misconception; P8, GPT-5).}
\begin{quote}\small
$z^2\sin(1/z)=z-(1/6)z^{-1}+(1/120)z^{-3}-\cdots$. For a contour
integral, only the constant term $z^0$ contributes. There is no constant
term, so the integral is 0 and $A=0$.
\end{quote}
This applies a faulty general rule. Contour integration selects the
$z^{-1}$ coefficient, not the constant coefficient.

\paragraph{E4 (wrong choice; P3, GPT-5-mini).}
\begin{quote}\small
Approximate $\sin(1/z)$ by its leading term $1/z$ on $|z|=1$, giving
$z^2\sin(1/z)\approx z$. Because $z$ is entire, Cauchy's theorem makes
its closed-contour integral 0. Therefore $A=0$.
\end{quote}
The response chooses a local leading-term approximation and then applies
Cauchy's theorem, a procedure that discards the residue-bearing term.

\paragraph{E5 (structural blindness; P3, GPT-5).}
\begin{quote}\small
Let $w=1/z$, so $z=1/w$ and $dz=-dw/w^2$. Keeping the same orientation,
rewrite the integral as $-\oint \sin(w)/w^4\,dw$. The $w^{-1}$
coefficient comes from the $w^3$ term of $\sin w$ and equals $-1/3!$.
Thus the integral is $-2\pi i(-1/6)=\pi i/3$, so $A=\pi/3$.
\end{quote}
The response fails to account jointly for the change of variable and the
orientation reversal of the contour, producing the wrong sign.

\section{Refusal Heuristic}
\label{app:refusal}

Refusal detection (\S\ref{sec:metric}) is implemented as a
case-insensitive regex over the generation's final output text. Before
matching, typographic apostrophes (U+2018, U+2019, U+02BC, and U+FF07)
are normalized to ASCII so identical wording cannot receive a different
label because of typography. A
response that matches any pattern below receives the terminal status
\textsf{Refusal}, independently of the legacy joint author label.

\begin{quote}\small\ttfamily
I don't (actually) know \quad
I can't recall \quad
I'm not (fully/entirely) sure \quad
I'm uncertain \quad
I don't remember \quad
I vaguely recall \quad
lacking that (crucial) fact/detail/knowledge \quad
I don't have the precise/exact \quad
I'll (make a) cautious guess \quad
I have no idea \quad
I'm not familiar with \quad
I'm not entirely certain \quad
I don't fully know/understand/recall \quad
cannot confidently/fully decide/answer/recall \quad
not fully/completely sure \quad
I'm (still) (a bit) unclear/confused \quad
I'm unable to \quad
I fail to recall \quad
don't (exactly) know how
\end{quote}

The same flag appears in the released JSONL
(\textsf{is\_refusal}, see Appendix~\ref{app:dataset}). The
heuristic is intentionally permissive. Any explicit
epistemic-refusal phrasing triggers the flag, even when the
response also commits to a concrete final answer. This
conservative choice errs on the side of excluding borderline
cases from TEGSR; downstream users who want a
stricter criterion can filter on
\textsf{is\_refusal == False AND human\_examination == 1}, and
users who want the looser original criterion can ignore the
flag.

\paragraph{Scope.} Across the 2{,}700 Tier-1 cells, the normalized regex flags
424 responses, all on class E2 (knowledge gap); 376 have legacy
$h=1$ and are therefore excluded from TEGSR, while 48 already have
$h=0$. The two-backend release contains 295 flagged cells (290 with
$h=1$ and five with $h=0$).
Unfiltered joint-label counts can be recovered by ignoring the flag.

\section{Prompt Templates}
\label{app:prompts}

\paragraph{Few-shot Generation Agent prompts.} Every reported \GA{}
condition uses few-shot prompts containing Q-A-E 3-tuples (Question,
erroneous Answer, Explanation of why this exemplifies the target class).
The base system prompt contains eight such triples, with two each for
E1--E3 and one each for E4--E5. The expanded prompt contains 25,
five per class. The supplementary release provides the complete
system-prompt bodies and every per-call user template, copied from the
experimental notebook without editorial correction. The base source and
runtime strings are identical. In the expanded GA block and its matching EA
definition block, however, the notebook's ordinary non-raw Python literals
interpret some LaTeX-like backslash sequences at runtime; the release
therefore also includes the exact implemented runtime strings, JSON-encoded
so that the control characters are recovered without ambiguity.

\begin{table}[h]
\centering
\small
\begin{tabular}{lccccc}
\toprule
\textbf{Condition} & \textbf{E1} & \textbf{E2} & \textbf{E3} & \textbf{E4} & \textbf{E5} \\
\midrule
P3 & B & B & B & B & B \\
P5 & B & B & X & X & B \\
P6 & X & X & X & X & X \\
P7 & X & X & X & X & X \\
\bottomrule
\end{tabular}
\caption{Base (B) versus expanded (X) example inventories by paper
condition and target class. Each entry applies to both the \GA{}
system prompt and the prompted-\EA{} definition block. B contains
$2/2/2/1/1$ E1--E5 examples; X contains five per class.}
\label{tab:prompt-inventory}
\end{table}

\paragraph{LLM-generated auxiliary subject context.}
The historical code calls these strings ``textbook summaries,'' but no
textbook file or retrieved passage was provided. Before the pipeline runs,
GPT-4o-mini received each Tier-1 question and a fixed prompt asking it to
name a relevant book and chapter and generate a chapter-style summary.
The resulting 20 strings were stored once, paired with questions
q000--q019 by list index, and reused across all three backends.
Book and chapter names are model suggestions rather than verified sources, so
we treat the strings as generated auxiliary context rather than authoritative
summaries or excerpts.

The code inserts a question's stored string verbatim: the \GA{}
template introduces it with a fixed sentence describing it as a
textbook chapter summary, and the prompted-\EA{} template labels it
\texttt{Textbook Summary:}. These historical labels are preserved
only for exact reproducibility. The supplementary release contains
the generation prompt, the stored strings with their hashes, and the
complete context-bearing \GA{} and \EA{} templates.

\begin{table}[h]
\centering
\small
\begin{tabular}{lcc}
\toprule
\textbf{Condition} & \textbf{\GA{} context} & \textbf{Prompted-\EA{} context} \\
\midrule
P3 & None & None \\
P4 & None & E1--E5 \\
P5 & E3/E4 & E2/E3/E4 \\
P6 & None & None \\
P7 & E1--E5 & E1--E5 \\
\bottomrule
\end{tabular}
\caption{Recipients of the stored LLM-generated auxiliary context. P3--P4
isolates EA-side insertion; P6--P7 changes both agents, and P5 varies by
class.}
\label{tab:context-recipients}
\end{table}

\paragraph{Added expanded E1 exemplar.}
The following constant-acceleration triple is absent from the base
prompt and present in the expanded prompt.
\begin{quote}
\begin{minipage}{0.94\linewidth}\small\ttfamily
Q3: A car accelerates from rest at $4\,\mathrm{m/s^2}$ for 3 seconds.
What is its final speed?\\
A3: Final speed = acceleration $\times$ time = $4 \times 3 = 11$ m/s.\\
E3: The student made a simple multiplication slip. $4 \times 3 = 12$,
not 11.
\end{minipage}
\end{quote}

The excerpts below illustrate the same prompt form in compact form.

\begin{quote}\small\ttfamily
\textbf{Few-shot prompt excerpt for class~1:}\\
**Mental Typo**: This error happens when a student is sloppy.\\
Q1: Twice Angie's age, plus 4, is 20. How old is Angie?\\
A1: Twice Angie's age is $20-4=16$. Angie is $16/2=7$.\\
E1: The student mistakenly calculates 7 instead of 8.\\
Q2: A roll of 25\,m wire weighs 5\,kg. How much does a 75\,m roll
weigh?\\
A2: The 75\,m roll is three times bigger... so the 75\,m roll weighs
$15\,\mathrm{kg} \times 3 = 45\,\mathrm{kg}$.\\
E2: The weight in the last step should be 5\,kg, not 15\,kg.\\
Generate an answer of error type~1 to the following question.
\end{quote}

The four excerpts below pair the paper's compact class definition with
a Q--A--E triple from the expanded experimental prompt. The triples are
lightly typeset and line-wrapped, but their questions, erroneous answers,
and error mechanisms are preserved. The complete source strings are in
the supplementary prompt files.

\paragraph{Class 2 (Knowledge gap).}
\begin{quote}\small\ttfamily
**Knowledge Gap**: Missing definition, term, or formula.\\
Q3: What does the p-value represent in a hypothesis test?\\
A3: The p-value is the probability.\\
E3: The student didn't know the meaning of p-value. The p-value is
the probability of observing data as extreme as the sample, assuming
the null hypothesis is true. It is not the probability that the null
hypothesis is true.
\end{quote}

\paragraph{Class 3 (Misconception).}
\begin{quote}\small\ttfamily
**Misconception**: A faulty mental model fitted to prior
experience.\\
Q3: The standard deviation of a dataset is 5. What is the variance?\\
A3: Variance $= \sqrt{5} = 2.24$.\\
E3: The student confused variance and standard deviation. Variance
is the square of the standard deviation, not the square root. The
correct variance is $5^2 = 25$.
\end{quote}

\paragraph{Class 4 (Wrong choice).}
\begin{quote}\small\ttfamily
**Wrong Choice**: Selection of the wrong solution procedure.\\
Q4: Find the probability of drawing an ace from a standard deck of
52 cards.\\
A4: Probability $= \frac{4!}{52!} \approx 7.7 \times 10^{-68}$.\\
E4: The student misapplied factorial notation, confusing
permutations/combinations with simple probability. The correct answer
is $\frac{4}{52} = \frac{1}{13} \approx 0.0769$.
\end{quote}

\paragraph{Class 5 (Structural blindness).}
\begin{quote}\small\ttfamily
**Structural Blindness**: Failure to model component interaction.\\
Q3: In a group of students, 30 like math, 20 like science, and 10
like both. How many students like only one subject?\\
A3: Total $= 30 + 20 = 50$. Only one subject $= 50 - 10 = 40$.\\
E3: The student added both groups and subtracted the intersection
once. ``Only one'' requires subtracting it twice, giving
$(30-10)+(20-10)=30$. The student did not properly parse the
overlapping components and their exclusion criteria.
\end{quote}

\paragraph{Prompted Examination Agent system message (verbatim).}
The standard substantive judge call uses the following system-message
wording. Line wrapping is typographical, while the Markdown markers and
original misspelling are retained.
\begin{quote}\small\ttfamily\raggedright
You are an expert in language analysis. Your task is to determine whether a
given text exhibits a specific type of error, based solely on the provided
definition of that error.\\
You will be given:\\
* Definition: A **definition of that error type**\\
* Question: A **question**\\
* Sample Answer: A **sample answer** to the question\\
* An **error type class** (a positive integer, e.g., 1, 2, 3…)\\
* A **text to evaluate**\\[3pt]
Your job is to:\\
1. Decide whether the text fits the error type as defined, based on the
provided definition, question and sample answer.\\
2. Return the error class number based on the text and your justification. If
the text does not match any error definition, return 0 and your
justificaiton.\\[3pt]
DO NOT RETURN ANYTHING ELSE.
\end{quote}
The base or expanded definition-and-example block fills the
\texttt{Definition:} field of the user message rather than the system message;
the remaining fields are the question, gold answer, requested class, and
candidate. Context-bearing calls (Table~\ref{tab:context-recipients}) use an
otherwise identical system message with the additional exact bullet
\texttt{* Textbook Summary: One **chapter summary** of a textbook}; this
historical label denotes the generated auxiliary context described above.
The original loop then sends the judge response through a second call with
system message \texttt{You're a useful agent on extracting information.} and
extracts the class numeral. The standard and context-bearing system messages,
both user templates, and the extraction messages are released verbatim under
\path{prompts/} and indexed with hashes in \path{prompts/README.md}. P8 uses
the fine-tuned classifier and therefore has no prompted-EA system message.
The quoted full message applies to Tier~1 and the headline GPT-5 Tier-2 P1/P3
runs. 

\section{Questions Used in the Provisional First-20 Pilot}
\label{app:questions}

The 20-question Tier-1 subset from TheoremQA
\citep{chen2023theoremqa} is used throughout
Tables~\ref{tab:overall}, \ref{tab:agent-acc-ga},
and \ref{tab:redo}. The exact first-20 list
is shown in Table~\ref{tab:first20-questions}; each question is paired
with each of the five target error classes, yielding 100
(question, class) cells per condition and backend. The subset is the
first 20 questions in the local ordering, selected by convenience to
limit verification cost; it was not chosen by power analysis,
probability sampling, or subject stratification.

\begin{table*}[t]
\centering
\small
\begin{tabular}{rllp{0.67\textwidth}}
\toprule
\# & Answer type & Gold answer & Question summary \\
\midrule
1 & integer & 11760 & Ordered partition of an 8-element set into 5 non-empty ordered subsets. \\
2 & float & 1.0 & Definite integral $\int_{-\infty}^{+\infty}\sin(3t)\sin(t/\pi)/t^2\,dt$. \\
3 & float & 1.42 & Solve $2x^3+e^x=10$ using the Newton--Raphson method. \\
4 & integer & 4200 & Ordered partition of a 7-element set into 4 non-empty ordered subsets. \\
5 & list & [0, 5] & Shortest path from node 0 to node 5 in a given undirected graph. \\
6 & list & [4, 1, 2, 0] & Shortest path from node 4 to node 0 in a second undirected graph. \\
7 & float & -1.047 & Complex contour integral $\int_{|z|=1} z^2\sin(1/z)\,dz$, reported as the real coefficient of $i$. \\
8 & float & 0.955 & Fluid-volume calculation from container weight, water density, and slug mass. \\
9 & list & [1, 1] & Sequential compactness of an integral family of bounded functions; identify the applicable theorem. \\
10 & integer & 0 & $x$-value for a two-equation linear system. \\
11 & float & 1.12 & Escape-speed calculation for a projectile leaving Earth, in units of $10^4$ m/s. \\
12 & bool & False & Whether the nonnegative quadrant is a subspace of $\mathbb{R}^2$. \\
13 & list & [0.333, 0.25] & Compare two line integrals of $xy\,dx$ along a line segment and a parabola. \\
14 & bool & False & Feasibility of a simple connected planar graph with 200 vertices and 400 faces. \\
15 & list & [4, 2] & Coordinate vector in a nonstandard basis of $\mathbb{R}^2$. \\
16 & integer & 50000 & Number of labeled forests on 10 vertices with 5 components under vertex-separation constraints. \\
17 & integer & 1 & Derivative of the inverse of $f(x)=x+\cos x$ at 1. \\
18 & bool & True & Whether a sequence lies in the image of the left-shift operator on infinite real sequences. \\
19 & float & 1.3733 & Covariance of exponentiated Brownian motion values $X(s)$ and $X(t)$. \\
20 & float & 1.094 & Maximum entropy rate of a random walk on a connected graph with 4 edges. \\
\bottomrule
\end{tabular}
\caption{The first 20 TheoremQA questions used in the provisional
Tier-1 pilot. Long problem statements are summarized here; the exact
verbatim prompts and gold answers are released in
the released Tier-1 question-and-answer subset.}
\label{tab:first20-questions}
\end{table*}

\section{Annotation Protocol for the 1{,}600-Example \EA{} Training Set}
\label{app:annotation}

\paragraph{Training examples.} The 1{,}600 (3-tuple, judgment)
examples used to fine-tune the BERT-base-uncased \EA{} are
human-annotated, sampled to balance the five target error classes
(approximately 320 examples per class).

\paragraph{Label space.} Each example carries a single binary
``human examination'' label, released in the CSVs. It is $1$ iff an
author judged that the response is both genuinely incorrect
\emph{and} an instance of the requested target class, and $0$
otherwise (a correct answer, or a wrong answer in the wrong class).
Deployed classifier inference uses the 3-tuple
\texttt{"Question: \{q\} Answer: \{a\} Error Class: \{c\}"}.
A retrospective audit of the archived training notebook found that its
serialization instead appended
\texttt{"Human Examination: \{h\}"} to this text while also using $h$ as
the classification target. This is target leakage. The notebook used a
cell-random 60/20/20 split, so its original held-out score would not repair the
leakage; we do not report that score. Appendix~\ref{app:p2-ea-replay} evaluates
the deployed saved artifact without supplying $h$ and records source-pool
overlap explicitly.
Annotators referred to the five class definitions and the few-shot
exemplars shown in Appendix~\ref{app:prompts}.

\paragraph{Annotator pool.} Labels on the 1{,}600-example
training set were produced by the authors during the iterative
pilot phase rather than by recruited external annotators. The submitted
Tier-1 outcome labels were assigned by one author who knew the requested
class and experimental condition; the labeling was not blinded. No IRB
approval was required because no external human-subjects data were
collected.

\paragraph{Label resolution.} Tier-1 labels were assigned in one pass.
There was no independent second annotation pass, double annotation, or
adjudication, and hence no measured inter-annotator agreement. We treat
the labels as provisional. The planned blinded independent study is
future work rather than part of the submitted procedure.

\section{Per-Pipeline Per-Class TEGSR}
\label{app:per-pipeline}

This section gives the per-class breakdown for the same Tier-1
sweep summarized by averages in main-paper Table~\ref{tab:overall}.
Tier-1 conditions with an \EA{} retry without a cap until operational
acceptance; P0/P2 instead store one draft. Every stored response is then
assigned the provisional one-pass label described in
Appendix~\ref{app:annotation}. Table~\ref{tab:agent-acc-ga} reports the post-hoc
outcome in Equation~\ref{eq:tegsr}, not \EA{} acceptance or
human--\EA{} agreement. An accepted false positive has score~0.

Table~\ref{tab:agent-acc-ga} shows two patterns. First, backend averages
over P1--P7 are close: $0.64$ for o3$+$GPT-4o and $0.65$ for each
GPT-5 backend. Second, E2 is the bottleneck under the conservative
refusal exclusion (backend--pipeline rates $0.00$--$0.35$). E5 remains
difficult on several configurations ($0.30$--$0.75$), including GPT-5
P8 ($0.60$) versus P3 ($0.70$). No backend or pipeline has a stable
point-estimate advantage across classes.

\paragraph{Context conditions P4--P7.} The per-class results clarify why
their aggregates are not uniform. Relative to P3, EA-only context in P4
changes E1--E5 by $-0.10/-0.05/0/+0.05/-0.15$ on o3$+$GPT-4o,
$-0.10/+0.05/+0.10/+0.05/-0.15$ on GPT-5, and
$-0.25/-0.10/+0.10/0/-0.15$ on GPT-5-mini. The common E5 decrease
contrasts with flat or improved E3/E4 values. Adding context to both agents
in P6$\to$P7 raises four classes on o3$+$GPT-4o
($+0.15/-0.05/+0.20/+0.05/+0.05$) and GPT-5-mini
($+0.10/+0.20/+0.05/0/+0.10$); GPT-5 improves E2/E3
($-0.05/+0.05/+0.10/-0.05/0$). P5 is not a clean per-class ablation. It
simultaneously changes example inventory and context recipient according to
the requested class. Its high E4 values ($0.95/0.80/1.00$) and low E5 values
($0.65/0.35/0.35$) are therefore descriptive, not attributable to one input.
Each class--backend value averages only 20 questions from the convenience
sample, so these patterns do not establish population-level class effects.

\begin{table*}[t]
\centering
\small
\begin{tabular}{llcccccc}
\toprule
\textbf{Pipeline} & \textbf{Backend} & E1 & E2 & E3 & E4 & E5 & \textbf{Avg} \\
\midrule
\multirow{3}{*}{P0} & o3+4o & \textbf{0.75} & 0.10 & 0.70 & 0.65 & \textbf{0.55} & 0.55 \\
 & GPT-5 & 0.70 & \textbf{0.20} & \textbf{0.85} & 0.90 & 0.50 & 0.63 \\
 & GPT-5-mini & 0.60 & 0.15 & \textbf{0.85} & \textbf{0.95} & 0.35 & 0.58 \\
\midrule
\multirow{3}{*}{P1} & o3+4o & 0.70 & 0.10 & 0.55 & 0.65 & 0.50 & 0.50 \\
 & GPT-5 & \textbf{0.95} & 0.10 & \textbf{1.00} & 0.90 & \textbf{0.55} & 0.70 \\
 & GPT-5-mini & \textbf{0.95} & \textbf{0.25} & 0.85 & \textbf{0.95} & 0.45 & 0.69 \\
\midrule
\multirow{3}{*}{P2} & o3+4o & 0.75 & 0.05 & 0.75 & \textbf{0.95} & \textbf{0.60} & 0.62 \\
 & GPT-5 & \textbf{0.90} & 0.00 & \textbf{0.95} & 0.75 & 0.50 & 0.62 \\
 & GPT-5-mini & 0.70 & \textbf{0.35} & 0.85 & 0.85 & 0.40 & 0.63 \\
\midrule
\multirow{3}{*}{P3} & o3+4o & 0.85 & 0.15 & 0.85 & \textbf{0.95} & \textbf{0.75} & 0.71 \\
 & GPT-5 & \textbf{0.95} & 0.00 & \textbf{0.90} & 0.90 & 0.70 & 0.69 \\
 & GPT-5-mini & 0.90 & \textbf{0.20} & 0.85 & 0.90 & 0.45 & 0.66 \\
\midrule
\multirow{3}{*}{P4} & o3+4o & 0.75 & \textbf{0.10} & 0.85 & \textbf{1.00} & \textbf{0.60} & 0.66 \\
 & GPT-5 & \textbf{0.85} & 0.05 & \textbf{1.00} & 0.95 & 0.55 & 0.68 \\
 & GPT-5-mini & 0.65 & \textbf{0.10} & 0.95 & 0.90 & 0.30 & 0.58 \\
\midrule
\multirow{3}{*}{P5} & o3+4o & 0.75 & 0.15 & 0.80 & 0.95 & \textbf{0.65} & 0.66 \\
 & GPT-5 & \textbf{0.90} & 0.15 & 0.85 & 0.80 & 0.35 & 0.61 \\
 & GPT-5-mini & 0.85 & \textbf{0.35} & \textbf{0.90} & \textbf{1.00} & 0.35 & 0.69 \\
\midrule
\multirow{3}{*}{P6} & o3+4o & 0.65 & 0.05 & 0.80 & \textbf{0.95} & \textbf{0.65} & 0.62 \\
 & GPT-5 & \textbf{0.90} & 0.05 & 0.80 & 0.85 & 0.45 & 0.61 \\
 & GPT-5-mini & 0.75 & \textbf{0.15} & \textbf{0.85} & 0.90 & 0.40 & 0.61 \\
\midrule
\multirow{3}{*}{P7} & o3+4o & 0.80 & 0.00 & \textbf{1.00} & \textbf{1.00} & \textbf{0.70} & 0.70 \\
 & GPT-5 & \textbf{0.85} & 0.10 & 0.90 & 0.80 & 0.45 & 0.62 \\
 & GPT-5-mini & \textbf{0.85} & \textbf{0.35} & 0.90 & 0.90 & 0.50 & 0.70 \\
\midrule
\multirow{3}{*}{P8} & o3+4o & 0.90 & 0.10 & 0.80 & 0.90 & 0.55 & 0.65 \\
 & GPT-5 & 0.85 & 0.15 & \textbf{0.85} & 0.85 & \textbf{0.60} & 0.66 \\
 & GPT-5-mini & \textbf{0.95} & \textbf{0.35} & 0.75 & \textbf{0.95} & 0.50 & 0.70 \\
\bottomrule
\end{tabular}
\caption{Provisional Tier-1 pilot per-class TEGSR on the first-20 TheoremQA convenience sweep, broken out by paper-label pipeline and backend. Single-author-verified, with unlimited \GA/\EA{} retries. Per-class peaks across backends are in bold.}
\label{tab:agent-acc-ga}
\end{table*}

\section{Per-Subject Tier-1 Breakdown}
\label{app:subject}

\S\ref{sec:framework} describes a common \GA/\EA{} protocol with no
subject-specific module. To describe how Tier-1 TEGSR varies by subject
under that common protocol, we re-bucket the same per-cell
labels that feed Table~\ref{tab:overall} by the subject of the
underlying TheoremQA question. The 20 questions in the Tier-1 sweep
fall into four buckets, namely combinatorics and graphs (Q1, Q4, Q5, Q6,
Q14, Q16), analysis (Q2, Q3, Q7, Q9, Q13, Q17, Q18), linear algebra
(Q10, Q12, Q15), and physics and probability (Q8, Q11, Q19, Q20).
These 6/7/3/4 counts are the realized composition of a convenience
sample, not a pre-specified stratification. The per-subject means are
therefore descriptive and are reported in Table~\ref{tab:subject}; the
``All'' column reproduces Table~\ref{tab:overall} as a sanity check
on the bucketing.

The table does not support a common ordering across subjects. The
easiest and hardest subject buckets vary across conditions. Within a
single (pipeline, backend) row, the subject spread can also exceed
the within-backend pipeline differences discussed in
\S\ref{sec:tier1-pipelines}. On GPT-5 P3, for example, the
per-subject range is $0.60$--$0.80$ ($\Delta\!=\!0.20$), whereas
the GPT-5 P1$\to$P3 headline gap is $-0.01$. Identical prompts and
configurations therefore establish a common cross-subject protocol.
They do not imply equal performance; the actual success rate varies by subject.

\begin{table*}[t]
\centering
\small
\setlength{\tabcolsep}{5pt}
\begin{tabular}{llccccc}
\toprule
\textbf{Pipeline} & \textbf{Backend} & \textbf{Comb./graph} & \textbf{Analysis} & \textbf{Linear alg.} & \textbf{Phys./prob.} & \textbf{All} \\
\midrule
P0 & o3+4o & 0.73 & 0.49 & 0.40 & 0.50 & 0.55 \\
 & GPT-5 & 0.73 & 0.54 & 0.73 & 0.55 & 0.63 \\
 & GPT-5-mini & 0.70 & 0.51 & 0.53 & 0.55 & 0.58 \\
\midrule
P1 & o3+4o & 0.47 & 0.43 & 0.47 & 0.70 & 0.50 \\
 & GPT-5 & 0.77 & 0.63 & 0.80 & 0.65 & 0.70 \\
 & GPT-5-mini & 0.77 & 0.63 & 0.60 & 0.75 & 0.69 \\
\midrule
P2 & o3+4o & 0.70 & 0.51 & 0.47 & 0.80 & 0.62 \\
 & GPT-5 & 0.77 & 0.57 & 0.60 & 0.50 & 0.62 \\
 & GPT-5-mini & 0.77 & 0.54 & 0.60 & 0.60 & 0.63 \\
\midrule
P3 & o3+4o & 0.73 & 0.74 & 0.53 & 0.75 & 0.71 \\
 & GPT-5 & 0.80 & 0.63 & 0.73 & 0.60 & 0.69 \\
 & GPT-5-mini & 0.77 & 0.60 & 0.73 & 0.55 & 0.66 \\
\midrule
P4 & o3+4o & 0.70 & 0.69 & 0.40 & 0.75 & 0.66 \\
 & GPT-5 & 0.73 & 0.66 & 0.67 & 0.65 & 0.68 \\
 & GPT-5-mini & 0.63 & 0.57 & 0.60 & 0.50 & 0.58 \\
\midrule
P5 & o3+4o & 0.70 & 0.66 & 0.40 & 0.80 & 0.66 \\
 & GPT-5 & 0.80 & 0.51 & 0.60 & 0.50 & 0.61 \\
 & GPT-5-mini & 0.87 & 0.66 & 0.67 & 0.50 & 0.69 \\
\midrule
P6 & o3+4o & 0.67 & 0.63 & 0.40 & 0.70 & 0.62 \\
 & GPT-5 & 0.73 & 0.51 & 0.67 & 0.55 & 0.61 \\
 & GPT-5-mini & 0.80 & 0.40 & 0.67 & 0.65 & 0.61 \\
\midrule
P7 & o3+4o & 0.73 & 0.71 & 0.53 & 0.75 & 0.70 \\
 & GPT-5 & 0.87 & 0.51 & 0.67 & 0.40 & 0.62 \\
 & GPT-5-mini & 0.77 & 0.66 & 0.67 & 0.70 & 0.70 \\
\midrule
P8 & o3+4o & 0.67 & 0.57 & 0.73 & 0.70 & 0.65 \\
 & GPT-5 & 0.77 & 0.49 & 0.80 & 0.70 & 0.66 \\
 & GPT-5-mini & 0.80 & 0.60 & 0.73 & 0.70 & 0.70 \\
\bottomrule
\end{tabular}
\caption{Descriptive per-subject TEGSR for the provisional Tier-1 first-20 convenience sweep, broken out by paper-label pipeline and backend. The sample was not subject-stratified: combinatorics/graph $n{=}30$ (6 questions $\times$ 5 classes), analysis $n{=}35$ (7 $\times$ 5), linear algebra $n{=}15$ (3 $\times$ 5), and physics/probability $n{=}20$ (4 $\times$ 5). The ``All'' column ($n{=}100$) matches Table~\ref{tab:overall}. Within-row variation describes this sample under the common protocol in \S\ref{sec:framework}.}
\label{tab:subject}
\end{table*}

\section{Retry Counts}
\label{app:redo}

Across all conditions with an \EA{}, on \EA{} rejection the \GA{}
regenerates a candidate; an acceptance ends the loop.
Table~\ref{tab:redo} reports the mean number of \GA{} attempts
needed before a candidate is accepted by the \EA{}. Thus a value of
1.00 means the first generation was accepted, and the number of
actual retries is the table value minus one. P2 has no \EA{} loop,
so it is included only as a reference row with value 1.00.

The table explains the cost behavior behind the backend comparison
in Table~\ref{tab:overall}. The o3$+$GPT-4o stack often needs
multiple attempts, especially for class~5; GPT-5-mini usually
reduces this to one or two attempts; GPT-5 reduces most cells to a
near-single-pass regime. The exception is class~5, where even GPT-5
still needs modest extra search in P1/P3/P6/P5. This supports the
paper's interpretation that structural blindness is not merely a
weak-model artifact but a persistent hard target.

\begin{table*}[!tp]
\centering
\small
\begin{tabular}{llcccccc}
\toprule
\textbf{Pipeline} & \textbf{Backend} & E1 & E2 & E3 & E4 & E5 & All \\
\midrule
\multirow{3}{*}{P1} & o3+4o & 1.45 & 3.45 & 2.00 & 2.10 & 3.25 & 2.45 \\
                   & GPT-5     & 1.10 & 1.10 & 1.05 & 1.25 & 1.90 & 1.28 \\
                   & GPT-5-mini & 1.25 & 1.30 & 1.20 & 1.35 & 1.55 & 1.33 \\
\multirow{3}{*}{P2} & o3+4o & 1.00 & 1.00 & 1.00 & 1.00 & 1.00 & 1.00 \\
                   & GPT-5     & 1.00 & 1.00 & 1.00 & 1.00 & 1.00 & 1.00 \\
                   & GPT-5-mini & 1.00 & 1.00 & 1.00 & 1.00 & 1.00 & 1.00 \\
\multirow{3}{*}{P3} & o3+4o & 2.50 & 3.30 & 1.70 & 1.85 & 6.10 & 3.09 \\
                   & GPT-5     & 1.10 & 1.00 & 1.05 & 1.15 & 1.35 & 1.13 \\
                   & GPT-5-mini & 1.10 & 1.10 & 1.25 & 1.80 & 1.30 & 1.31 \\
\multirow{3}{*}{P4} & o3+4o & 2.30 & 2.85 & 1.70 & 1.60 & 5.45 & 2.78 \\
                   & GPT-5     & 1.10 & 1.00 & 1.10 & 1.65 & 1.20 & 1.21 \\
                   & GPT-5-mini & 1.10 & 1.15 & 1.40 & 1.85 & 2.25 & 1.55 \\
\multirow{3}{*}{P5} & o3+4o & 2.20 & 2.05 & 3.20 & 2.70 & 3.00 & 2.63 \\
                   & GPT-5     & 1.05 & 1.00 & 1.20 & 1.10 & 1.40 & 1.15 \\
                   & GPT-5-mini & 1.20 & 1.20 & 1.40 & 1.35 & 1.40 & 1.31 \\
\multirow{3}{*}{P6} & o3+4o & 1.75 & 3.95 & 1.85 & 1.55 & 4.35 & 2.69 \\
                   & GPT-5     & 1.00 & 1.15 & 1.05 & 1.20 & 1.40 & 1.16 \\
                   & GPT-5-mini & 1.10 & 1.45 & 1.30 & 1.35 & 1.50 & 1.34 \\
\multirow{3}{*}{P7} & o3+4o & 2.50 & 4.45 & 2.55 & 2.50 & 4.45 & 3.29 \\
                   & GPT-5     & 1.10 & 1.00 & 1.00 & 1.15 & 1.25 & 1.10 \\
                   & GPT-5-mini & 1.25 & 1.20 & 1.80 & 1.35 & 1.45 & 1.41 \\
\multirow{3}{*}{P8} & o3+4o & 1.05 & 1.00 & 1.00 & 1.00 & 1.25 & 1.06 \\
                   & GPT-5     & 1.25 & 1.00 & 1.90 & 1.40 & 1.15 & 1.34 \\
                   & GPT-5-mini & 1.05 & 1.00 & 1.25 & 1.15 & 1.10 & 1.11 \\
\bottomrule
\end{tabular}
\caption{Mean number of \GA{} attempts per accepted generation
(\emph{attempts}, not retries; actual retries equal attempts minus
one). Pipeline~2 has no examination loop and trivially equals~1.
Class~5 (structural blindness) is the most retry-intensive class
across pipelines and backends, except for P8 where the BERT
classifier is near-single-pass on all backends, with its largest
retry pocket on class~3 (misconception) for GPT-5 and GPT-5-mini
and on class~5 for o3$+$GPT-4o.}
\label{tab:redo}
\end{table*}

\paragraph{How often does the loop need more than five attempts?}
Tier-1 uses unlimited retries, but Tier-2 (\S\ref{sec:tier2})
deploys a 5-attempt cap. Table~\ref{tab:redo-gt5} counts, for each
(pipeline, backend, error class), how many of the 20 \textsf{(question,
target class)} cells required more than five total attempts to be
accepted---equivalently, \textsf{redo\_count}$\geq5$. Of 2{,}100
Tier-1 cells with an \EA{} loop (7 pipelines $\times$ 3 backends
$\times$ 100 cells), 75 (3.6\%) cross this threshold; 64 of those
(85\%) are on the o3$+$GPT-4o backend, concentrated on
E5 (structural blindness) and secondarily E2 (knowledge gap). On
GPT-5 and GPT-5-mini the cap rarely binds; in particular P3 on GPT-5
has \emph{zero} cells requiring more than five attempts, supporting the deployment
choice of a 5-attempt cap for the Tier-2 scale runs.

\begin{table*}[!tp]
\centering
\small
\setlength{\tabcolsep}{5pt}
\begin{tabular}{llcccccc}
\toprule
\textbf{Pipeline} & \textbf{Backend} & \textbf{E1} & \textbf{E2} & \textbf{E3} & \textbf{E4} & \textbf{E5} & \textbf{All} \\
\midrule
\multirow{3}{*}{P1} & o3+4o & 0 & 1 & 0 & 0 & 4 & 5/100 (5\%) \\
 & GPT-5 & 0 & 0 & 0 & 0 & 1 & 1/100 (1\%) \\
 & GPT-5-mini & 0 & 0 & 0 & 0 & 0 & 0/100 \\
\midrule
\multirow{3}{*}{P3} & o3+4o & 2 & 4 & 1 & 1 & 8 & 16/100 (16\%) \\
 & GPT-5 & 0 & 0 & 0 & 0 & 0 & 0/100 \\
 & GPT-5-mini & 0 & 0 & 0 & 1 & 0 & 1/100 (1\%) \\
\midrule
\multirow{3}{*}{P4} & o3+4o & 1 & 3 & 1 & 0 & 5 & 10/100 (10\%) \\
 & GPT-5 & 0 & 0 & 0 & 1 & 0 & 1/100 (1\%) \\
 & GPT-5-mini & 0 & 0 & 0 & 2 & 2 & 4/100 (4\%) \\
\midrule
\multirow{3}{*}{P5} & o3+4o & 0 & 1 & 2 & 2 & 5 & 10/100 (10\%) \\
 & GPT-5 & 0 & 0 & 0 & 0 & 0 & 0/100 \\
 & GPT-5-mini & 0 & 0 & 0 & 0 & 0 & 0/100 \\
\midrule
\multirow{3}{*}{P6} & o3+4o & 0 & 3 & 0 & 0 & 6 & 9/100 (9\%) \\
 & GPT-5 & 0 & 0 & 0 & 0 & 0 & 0/100 \\
 & GPT-5-mini & 0 & 0 & 0 & 0 & 0 & 0/100 \\
\midrule
\multirow{3}{*}{P7} & o3+4o & 1 & 5 & 1 & 2 & 5 & 14/100 (14\%) \\
 & GPT-5 & 0 & 0 & 0 & 0 & 0 & 0/100 \\
 & GPT-5-mini & 0 & 0 & 1 & 0 & 1 & 2/100 (2\%) \\
\midrule
\multirow{3}{*}{P8} & o3+4o & 0 & 0 & 0 & 0 & 0 & 0/100 \\
 & GPT-5 & 0 & 0 & 1 & 0 & 0 & 1/100 (1\%) \\
 & GPT-5-mini & 0 & 0 & 1 & 0 & 0 & 1/100 (1\%) \\
\bottomrule
\end{tabular}
\caption{Provisional Tier-1 pilot attempts$>$5 counts per (pipeline, backend, error class) on the first-20 TheoremQA convenience sweep. Per-class cells are counts out of $n{=}20$; the ``All'' column is the per-(pipeline, backend) total out of $n{=}100$ with per-row percentage. Pipelines~P0 and P2 are omitted because they have no \EA{} loop. Because retries equal attempts minus one, attempts$>$5 is exactly \textsf{redo\_count}$\geq5$. The tail concentrates on the o3$+$GPT-4o backend (85\% of these cells) and on classes E5 and E2; on GPT-5 and GPT-5-mini the 5-attempt cap rarely binds.}
\label{tab:redo-gt5}
\end{table*}

\section{Fixed-Output \EA{} Comparison}
\label{app:p2-ea-replay}

To separate selector behavior from regeneration, we held fixed the same 300
paper-P2 candidates and their prior author joint-success labels, then applied
the same conservative refusal exclusion used for headline TEGSR. Each
backend's original prompted \EA{} judged its 100 candidates once; the saved P8
classifier was then applied locally to all 300 using its deployed question,
candidate, and requested-class input. The P8 replay required no API call,
generation, or new annotation. Table~\ref{tab:p2-ea-comparison} reports
coverage, precision, recall, and retained-valid yield. Precision alone is
insufficient because a selective verifier can improve it by discarding valid
candidates.

\begin{table}[H]
\centering
\small
\setlength{\tabcolsep}{2pt}
\begin{tabular}{llrrrr}
\toprule
Selector & Backend & Cov. & Prec. & Recall & Yield \\
\midrule
Prompted & o3$+$GPT-4o & 0.38 & 0.66 & 0.40 & 0.25 \\
 & GPT-5 & 0.92 & 0.65 & 0.97 & 0.60 \\
 & GPT-5-mini & 0.92 & 0.61 & 0.89 & 0.56 \\
\midrule
P8 & o3$+$GPT-4o & 0.98 & 0.62 & 0.98 & 0.61 \\
 & GPT-5 & 0.92 & 0.64 & 0.95 & 0.59 \\
 & GPT-5-mini & 0.93 & 0.68 & 1.00 & 0.63 \\
\bottomrule
\end{tabular}
\caption{Fixed-output selector diagnostics on the same 100 paper-P2 candidates per backend. Precision and recall use the provisional legacy author joint label after the conservative refusal exclusion. The P8 rows describe the deployed saved artifact, not clean held-out performance. Its archived training text contains the label, and the GPT-5/GPT-5-mini candidates belong to its source pool.}
\label{tab:p2-ea-comparison}
\end{table}

P8 accepts $0.92$--$0.98$ of the fixed candidates. Its high recall
($0.95$--$1.00$) therefore coexists with low invalid-output specificity
($0.03/0.13/0.19$ for o3$+$GPT-4o / GPT-5 / GPT-5-mini). These values are
artifact diagnostics rather than a held-out estimate. The training audit found
label leakage, and exact text matching places all 200 GPT-5/GPT-5-mini P2
candidates in P8's QAEH source pool (125 train, 35 validation, 40 test); only
the 100 o3$+$GPT-4o candidates are outside it.

The prompted-EA append-only artifact preserves all 300 raw responses, prompt
hashes, returned model versions, timestamps, and token usage. Its companion
analysis reports question-clustered intervals. The P8 artifact preserves every
binary decision, acceptance probability, weight hash, and reconstructed split
membership. Neither analysis measures run-to-run generation variability.

\section{Three-Agent Cascade Ablation}
\label{app:ablation-cascade}

To test whether adding more judges to the \GA/\EA{} pipeline would
help, we run a small ablation in which a third agent (an independent
fresh \EA{} call using the same prompt and the same model, GPT-5)
re-judges each of the 100 accepted P3 outputs. The
question is whether a third judge would reject cells the first
judge accepted. If so, end-to-end acceptance in a 3-agent cascade
would drop proportionally; if not, the third judge contributes no
new information.

\begin{table}[!tbp]
\centering
\small
\begin{tabular}{lrr}
\toprule
\textbf{Outcome} & \textbf{Author-valid} & \textbf{Author-invalid} \\
\midrule
EA1 accept & 69 & 31 \\
EA2 accept & 67 & 29 \\
EA2 reject & 2 & 2 \\
\bottomrule
\end{tabular}
\caption{Three-agent cascade diagnostic on P3 GPT-5. A fresh prompted
\EA{} retains 96 of 100 EA1-accepted cells; the columns join those decisions
with the provisional refusal-aware author label.}
\label{tab:ablation-cascade}
\end{table}

The second judge's retention agreement is $0.96$, but agreement alone is
neither precision nor recall. Against the existing author labels, its precision
is $67/96=0.70$, recall is $67/69=0.97$, and retained-valid yield is $0.67$.
Before that judge, the 100 EA1-accepted candidates have author-valid precision
$0.69$, recall $1.00$, and yield $0.69$. Thus the second judge rejects two
author-invalid and two author-valid candidates. The four rejections occur in
requested classes E1, E3, and E4, not all five classes. Because the run stores
no replacement generations after EA2 rejection, it cannot estimate the TEGSR
of a complete three-agent retry pipeline and does not establish a benefit from
adding the third judge.

\section{Structured Comparison with Generation and Evaluation Precedents}
\label{app:method-comparison}

Table~\ref{tab:method-comparison} separates task definitions from mechanisms.
The first three rows are direct educational-error generators; the next two are
general evaluation or refinement patterns that motivate the use of a judge and
feedback. The final row describes our evaluated conditions. Because inputs,
outputs, ground truth, and evaluation sets differ, the reported values support
within-paper comparisons only.

\begin{table*}[!t]
\centering
\scriptsize
\setlength{\tabcolsep}{2.3pt}
\renewcommand{\arraystretch}{1.18}
\newcolumntype{P}[1]{>{\raggedright\arraybackslash}p{#1}}
\begin{tabular}{@{}P{0.09\textwidth}P{0.14\textwidth}P{0.115\textwidth}P{0.135\textwidth}P{0.13\textwidth}P{0.145\textwidth}P{0.125\textwidth}@{}}
\toprule
\textbf{Work} & \textbf{Conditioning signal} & \textbf{Output format} & \textbf{Evaluator / selector} & \textbf{Feedback mechanism} & \textbf{Success criterion} & \textbf{Reported metric} \\
\midrule
\citet{scarlatos2024overgenerate}
& Math-MCQ stem, key, and key explanation
& Three ranked distractor options
& DPO ranker predicts relative student selection
& No test-time revision: generate 10, retain top 3
& Selected options overlap teacher distractors, favoring frequently chosen ones
& Partial, exact, proportional, and weighted-proportional match; rank accuracy \\

\citet{parikh2025lookalike}
& Math MCQ (optionally key, explanation, and tags) plus an error or distractor
& A paired distractor or free-text error description
& Exact match for distractors; GPT-4o-mini equivalence judge for errors
& Training-time mined negative preferences; alternating SFT/DPO
& Output is consistent with the paired teacher error or distractor
& 51.6\% distractor / 57.2\% error vs. DiVERT 45.6\% / 47.7\% \\

\citet{ross2025mistakes}
& Question and correct answer; inferred or supplied misconception
& Misconception, faulty reasoning trace, and wrong answer
& Cycle check; GPT-4o-mini equivalence judge for distractor evaluation
& Inner cycle filter/weight; outer iterative fine-tuning
& Misconception regenerates the same wrong answer; downstream output matches an expert label
& Simulation accuracy; misconception MAP@25; distractor precision \\
\midrule

\citet{zheng2023judging}
& Prompt plus one or two candidate chat responses
& Scalar score or pairwise preference, usually with rationale
& Strong LLM judge calibrated against human votes
& None: post-hoc evaluation, not regeneration
& Judge choice agrees with human preference
& Judge--human agreement (over 80\% for strong judges); MT-Bench scores / win rates \\

\citet{madaan2023self}
& Task input plus the model's current draft
& Iteratively revised task response
& The same LLM supplies feedback and revision
& Natural-language feedback conditions the next iteration
& Refined response improves task-specific quality
& Human preference and task-specific automatic metrics ($\sim$20\% absolute mean gain) \\
\midrule

\textbf{This work}
& Question, target class, and few-shot examples; P1--P8 also receive the gold final answer, but no condition receives a gold trace
& Open-ended worked incorrect response
& Prompted-LLM or BERT \EA{} selects; one author verifies provisional Tier-1 outcomes
& On rejection, \GA{} regenerates; P3--P7 also see the latest failure
& Incorrect final answer, requested error class, and no refusal
& Provisional single-author-verified TEGSR with exact question-cluster intervals \\
\bottomrule
\end{tabular}
\caption{Structured comparison with direct educational-error generators and
general evaluation/refinement patterns. Each row reports the success criterion
and metric of its own task and test set; the values are therefore not directly
comparable and do not establish cross-task state-of-the-art superiority.}
\label{tab:method-comparison}
\end{table*}

\section{Released Dataset Schema and Counts}
\label{app:dataset}

The supplementary release contains 1{,}800 records over the
TheoremQA main phase (20 questions $\times$ 5 error classes $\times$
9 pipelines $\times$ 2 backends), packaged as JSONL. The full
per-record schema and datasheet are included with the release.
It contains 1{,}166 TEGSR successes, 295 refusals, and 339 unresolved
non-refusal legacy negatives.

\paragraph{Per-record fields.} Each JSONL line contains the source
question and gold answer (verbatim from TheoremQA), the target
error class (1--5) and its human-readable name, the pipeline and
backend identifier, the generating GA's final stored response,
the \EA{}'s judgment and operational acceptance flag (null for
P0/P2), the number of retries before acceptance or termination, a
legacy binary \textsf{human\_examination} label, and the separate
\textsf{is\_refusal} status (\S\ref{sec:metric}). The release also
derives nullable \textsf{answer\_correct}, \textsf{realized\_class},
and \textsf{terminal\_outcome} fields, plus the complete boolean
\textsf{tegsr\_success}. A refusal resolves to \textsf{Refusal} and
TEGSR~0. A non-refusal legacy positive resolves to
\textsf{answer\_correct=False}, realized class equal to the target,
\textsf{WrongAns--TargetClass}, and TEGSR~1. For a non-refusal
legacy negative, the three decomposed fields remain null because the
source bit cannot distinguish \textsf{CorrectAns--ClassNA} from
\textsf{WrongAns--OtherClass}; TEGSR is nevertheless exactly~0.
Operational \textsf{ea\_accepted} is not part of the TEGSR equation.

\paragraph{Availability, license, and attribution.} Upon
publication, the main-phase records and the fine-tuned
BERT-base-uncased \EA{} weights from pipeline~P8 will be released
under CC BY-SA 4.0; the source questions are drawn from
TheoremQA~\citep{chen2023theoremqa} (MIT licensed), acknowledged in
the upstream attribution file. The data and
code are available at
\url{https://github.com/junli-cuny/taxonomy-targeted-error-generation}.

\paragraph{Privacy and content.} Because this study uses public questions
and synthetic outputs rather than authentic student data, the release contains
no personally identifying information. Every record is a synthetically
generated solution to a publicly released TheoremQA science question,
and no real student data was collected or stored. Each released record
carries the \textsf{human\_examination} verification label described
above, and that author review surfaced no offensive content,
consistent with the technical math and science domain of the source
questions.

\section{Compute and Reproducibility}
\label{app:compute}

\paragraph{Generation backends.} All \GA{} and prompted-\EA{}
inferences use OpenAI's commercial API. The backends are an OpenAI
o3$+$GPT-4o mixed configuration with \GA{} = OpenAI o3 and \EA{} = GPT-4o,
GPT-5, and GPT-5-mini.
Generations were collected in an offline batch regime,
decoupling pipeline depth from end-to-end latency at downstream
deployment time. Per-pipeline cost was a binding constraint and
influenced the decision to cap retries at five attempts and to
restrict the per-pipeline sweep to the first 20 questions of the
TheoremQA subset.

\paragraph{Fine-tuned \EA{}.} The BERT-base-uncased classifier in P8
is the standard \textsf{bert-base-uncased} model (12 layers,
768-dimensional hidden states, 12 attention heads; vocabulary size
30{,}522) configured for sequence classification with two output
labels. Fine-tuning was carried out on a single Google Colab GPU
using the HuggingFace \textsf{transformers} \textsf{Trainer}
(version 4.56.1) for 3 epochs, with
per-device train batch size 16, per-device eval batch size 64,
500 warmup steps, weight decay $0.01$, AdamW optimizer with the
HuggingFace default learning rate ($5 \times 10^{-5}$) and linear
schedule, and the HuggingFace \textsf{Trainer}'s default seed. The 1{,}600-example
training set was split 60/20/20 into train/validation/test
(960/320/320) using \texttt{sklearn}'s
\texttt{train\_test\_split} (\texttt{random\_state=42}).
Wall-clock fine-tuning time was on the order of minutes given
the small training-set size.

\paragraph{API usage.} Runs were collected offline; the q20+ scale
extensions additionally use the five-attempt cap described in
\S\ref{sec:framework}, whereas Tier-1 is uncapped. The JSON logs store per-call token
counts for the o3$+$GPT-4o, GPT-5, and GPT-5-mini runs. The
o3$+$GPT-4o backend was only run on the Tier-1 first-20 sweep,
whereas GPT-5 and GPT-5-mini also have q20+ 5-cap extensions. The
reported Tier-2 evidence is narrower. GPT-5 P1/P3/P8 cover q020--q199. Other
q20+ artifacts enter only this aggregate cost inventory and are not treated as Tier-2 validity
evidence.
Table~\ref{tab:tokencost} reports average per-cell token
consumption, \$-cost, and end-to-end latency by backend and
pipeline. Token counts include input, output, and
(billed-as-output) reasoning tokens across every retry, with
costs computed at OpenAI's public pricing as of 2026-05 (o3: input
\$$2.00$/$1$M, output \$$8.00$/$1$M; GPT-4o: input \$$2.50$/$1$M,
output \$$10$/$1$M; GPT-5: input \$$1.25$/$1$M, output \$$10$/$1$M;
GPT-5-mini: input \$$0.25$/$1$M, output \$$2$/$1$M; reasoning tokens
billed at the output rate).
P8's EA call is the fine-tuned BERT classifier and is treated as
free; only the GA call contributes to its API cost.

\paragraph{Sampling parameters.} All LLM calls use the API
provider's default temperature unless otherwise noted, with
top-$p$ left at default. The offline batch regime caches
generations so that downstream measurement is reproducible without
re-querying the APIs.

\paragraph{Run counts and variance.} Each
\textsf{(question, target class)} cell in Tables~\ref{tab:overall},
\ref{tab:agent-acc-ga}, and \ref{tab:redo}
is a single stochastic draw at temperature~1; we did not collect
multiple completions per cell to average over per-cell variance.
Cross-question variance is built into the means (20 questions
$\times$ 5 classes per pipeline), but the absolute TEGSR values
should be interpreted as estimates rather than calibrated point
values, as noted in the Limitations section.

\paragraph{Random seeds.} The train/validation/test split uses
\texttt{random\_state=42}. The BERT-\EA{} fine-tune itself relies
on the HuggingFace \textsf{Trainer}'s default seed; we did not set
\texttt{seed} explicitly in the \textsf{TrainingArguments}, so
replications should expect minor run-to-run variation in absolute
TEGSR values of order a few percentage points, consistent with
the single-sample-per-cell disclosure in the Limitations section.

\begin{table*}[t]
\centering
\small
\setlength{\tabcolsep}{4pt}
\begin{tabular}[t]{llrrr}
\toprule
\textbf{Backend} & \textbf{Pipe} & \textbf{Token} & \textbf{\$} & \textbf{Latency} \\
\midrule
o3+4o & P0 & 4,357 & \$0.024 & 19.4 \\
o3+4o & P1 & 19,156 & \$0.119 & 62.1 \\
o3+4o & P2 & 11,010 & \$0.074 & 45.4 \\
o3+4o & P3 & 41,608 & \$0.244 & 146.0 \\
o3+4o & P4 & 38,850 & \$0.220 & 135.0 \\
o3+4o & P5 & 44,321 & \$0.233 & 87.5 \\
o3+4o & P6 & 47,033 & \$0.237 & 133.5 \\
o3+4o & P7 & 66,911 & \$0.315 & 147.8 \\
o3+4o & P8 & 4,570 & \$0.022 & 18.6 \\
\midrule
gpt-5 & P0 & 6,286 & \$0.054 & 37.8 \\
gpt-5 & P1 & 24,670 & \$0.176 & 118.6 \\
gpt-5 & P2 & 5,856 & \$0.048 & 34.8 \\
gpt-5 & P3 & 22,757 & \$0.164 & 110.8 \\
gpt-5 & P4 & 14,643 & \$0.089 & 92.0 \\
\bottomrule
\end{tabular}\hspace{2.5em}
\begin{tabular}[t]{llrrr}
\toprule
\textbf{Backend} & \textbf{Pipe} & \textbf{Token} & \textbf{\$} & \textbf{Latency} \\
\midrule
gpt-5 & P5 & 16,501 & \$0.089 & 60.2 \\
gpt-5 & P6 & 16,469 & \$0.093 & 71.2 \\
gpt-5 & P7 & 19,847 & \$0.089 & 69.1 \\
gpt-5 & P8 & 7,864 & \$0.066 & 37.7 \\
\midrule
gpt-5-mini & P0 & 4,461 & \$0.007 & 26.1 \\
gpt-5-mini & P1 & 13,415 & \$0.015 & 49.8 \\
gpt-5-mini & P2 & 4,851 & \$0.007 & 23.4 \\
gpt-5-mini & P3 & 14,440 & \$0.017 & 51.4 \\
gpt-5-mini & P4 & 15,950 & \$0.017 & 54.0 \\
gpt-5-mini & P5 & 15,946 & \$0.014 & 56.2 \\
gpt-5-mini & P6 & 14,798 & \$0.014 & 42.9 \\
gpt-5-mini & P7 & 22,720 & \$0.017 & 53.4 \\
gpt-5-mini & P8 & 5,929 & \$0.007 & 24.4 \\
\bottomrule
\end{tabular}
\caption{Average per-cell token consumption, \$-cost, and end-to-end latency by backend and pipeline, aggregated over all available logged cells. ``tok./cell'' sums GA and EA input, output, and billed-as-output reasoning tokens across every retry. ``\$/cell'' uses OpenAI public pricing as of 2026-05 (o3 input \$2.00/1M, output \$8.00/1M; GPT-4o input \$2.50/1M, output \$10/1M; GPT-5 input \$1.25/1M, output \$10/1M; GPT-5-mini input \$0.25/1M, output \$2/1M). P8's \EA{} is a local BERT classifier and contributes zero API cost.}
\label{tab:tokencost}
\end{table*}

\section{Related Work}
\label{sec:related}

\paragraph{LLM reasoning evaluation.}
Chain-of-thought prompting \citep{wei2022chain}, process-reward
modeling \citep{lightman2024lets,uesato2022solving}, and
self-consistency decoding \citep{wang2023selfconsistency} have
established that intermediate reasoning quality is partially
decoupled from final answer accuracy. Divergent chains can reach
the same answer, and stepwise quality and final correctness can
be scored separately. Probing approaches (faithfulness tests
\citep{turpin2023language} and perturbation studies
\citep{mirzadeh2024gsmsymbolic}) ask whether models truly compute
the intermediate steps they verbalize. Our framing instead studies
\emph{controlled solution-side error generation}, which constructs a new
worked response conditioned on a target error class rather than modifying
the problem input or editing a supplied reasoning trace.

\paragraph{Educational error analysis and generation.}
Educational diagnosis is longstanding
\citep{brown1978diagnostic,vanlehn1990mind}; recent work classifies existing
responses in mathematics \citep{sun2025error,otero2024misconceptions},
programming \citep{oli2024gaps}, physics \citep{kortemeyer2023toward}, and
science \citep{wang2023scibench}. Direct generation is also established.
\citet{macina2023mathdial} simulate common errors in tutoring dialogues, and
\citet{sonkar2024cognitive} train algebra cognitive student models. For math
MCQs, \citet{feng2024distractor} generate distractors,
\citet{scarlatos2024overgenerate} overgenerate and rank them by predicted
student selection, and \citet{parikh2025lookalike} target error--distractor
consistency. \citet{ross2025mistakes} use cycle consistency among
misconceptions, reasoning, and wrong answers; \citet{zengaffinen2026incorrect}
find that correct-solution grounding improves alignment with human-authored
distractors. We therefore do not claim the first educational-error generator.
We instead evaluate open-ended worked responses conditioned on five fixed
cross-subject error classes under a joint incorrectness, class-fidelity, and
non-refusal success criterion.

\paragraph{LLMs as judges and multi-agent evaluation.}
LLM-as-judge methods \citep{zheng2023judging} and self-refinement
loops \citep{madaan2023self} have demonstrated that auxiliary models
can evaluate and revise primary-model output. Our task-specific
\GA/\EA{} loop instantiates these established patterns.
Table~\ref{tab:method-comparison} (Appendix~\ref{app:method-comparison})
compares conditioning, outputs, evaluators, feedback, success criteria, and
metrics; values are not cross-task comparable. Our two-agent setup is narrower
than general-purpose orchestration
\citep{wu2024autogen,hong2024metagpt}.

\paragraph{Synthetic data and error datasets.}
Synthetic data construction has become standard for training
mathematical reasoners \citep{yu2024metamath,luo2023wizardmath} and
code models \citep{luo2023wizardcoder}. Broad cross-subject datasets with
mechanism-level error labels remain limited. Benchmarks such as GSM8K
\citep{cobbe2021training}, TheoremQA \citep{chen2023theoremqa}, and
APPS \citep{hendrycks2021apps} provide only question--answer pairs
with no annotation of how a wrong answer would fail, and
process-reward datasets such as PRM800K
\citep{lightman2024lets} label reasoning steps as correct or
incorrect without identifying the underlying cognitive error type.
Our task and protocol support construction of datasets with controllable
error-class labels.

\end{document}